\documentclass[trsc,nonblindrev]{informs3} 

\OneAndAHalfSpacedXI 

\usepackage{natbib}
 \bibpunct[, ]{(}{)}{;}{a}{}{,}%
 \def\bibfont{\small}%
\usepackage{xcolor}
\usepackage{bm}
\usepackage{color}
\usepackage{amsmath}
\usepackage{booktabs}
\usepackage{tabularx}
\usepackage{threeparttable}
\usepackage{multicol}
\usepackage{multirow}
\usepackage{setspace}
\usepackage{graphicx}
\usepackage{subfigure}
\usepackage{algorithm} 
\usepackage{algorithmic}
\usepackage{authblk}
\usepackage{makecell}
\usepackage[section]{placeins}
\allowdisplaybreaks[4]
\usepackage[colorlinks=true, allcolors=blue]{hyperref}

\TheoremsNumberedThrough     

\EquationsNumberedThrough    

\begin{document}





\TITLE{Beyond the Next Port: A Multi-Task Transformer for Forecasting Future Voyage Segment Durations}

\ARTICLEAUTHORS{%
\AUTHOR{Nairui Liu, Fang He}
\AFF{Department of Industrial Engineering, Tsinghua University, Beijing 100084, P.R. China}
\AUTHOR{Xindi Tang}
\AFF{School of Management Science and Engineering, Central University of Finance and Economics, Beijing 100081, P.R. China}
\AUTHOR{Yineng Wang}
\AFF{Department of Logistics and Maritime Studies, The Hong Kong Polytechnic University, Hong Kong, P.R. China}
} 

\ABSTRACT{%
Accurate forecasts of segment-level sailing durations are fundamental to enhancing maritime schedule reliability and optimizing long-term port operations. However, conventional estimated time of arrival (ETA) models are primarily designed for the immediate next port of call and rely heavily on real-time automatic identification system (AIS) data, which is inherently unavailable for future voyage segments. To address this gap, the study reformulates future-port ETA prediction as a segment-level time-series forecasting problem. We develop a transformer-based architecture that integrates historical sailing durations, destination port congestion proxies, and static vessel descriptors. The proposed framework employs a causally masked attention mechanism to capture long-range temporal dependencies and a multi-task learning head to jointly predict segment sailing durations and port congestion states, leveraging shared latent signals to mitigate high uncertainty. Evaluation on a real-world global dataset from 2021 demonstrates the proposed model consistently outperforms a comprehensive suite of competitive baselines. The result shows a relative reduction of \textcolor{black}{4.70\%} in mean absolute error (MAE), 4.95\% in mean absolute percentage error (MAPE) \textcolor{black}{and 2.59\% in root mean squared error (RMSE)} compared with sequential deep learning models. The relative reductions compared with gradient boosting machines are \textcolor{black}{7.03\%} in MAE, \textcolor{black}{39.49\%} in MAPE \textcolor{black}{and 4.37\% in RMSE}. The case study conducted on one major destination port further illustrates the model's superior accuracy.
}%


\KEYWORDS{Maritime logistics, Sailing duration prediction, Future voyage segments, Multi-task learning, Transformer-based model, Segment-level forecasting}


\maketitle

%

\section{Introduction}

The maritime industry serves as a cornerstone of the global logistics framework and a primary conduit for international commerce \citep{chu2024vessel, wang2023innovative}. According to \textit{Review of Maritime Transport 2024} published by the United Nations Conference on Trade and Development, maritime transport facilitates approximately 80\% of global trade by volume. In 2023, global seaborne trade volumes ascended to 12.292 billion tons, representing an estimated 62.037 trillion ton-miles, thereby solidifying its status as the main artery of global trade \citep{unctad}. 

Operational efficiency in this domain is fundamentally predicated on the precision of estimated time of arrival (ETA) predictions. Accurate ETA forecasts empower port managers to optimize berth allocation, crane assignments, and yard space management, while facilitating the proactive synchronization of trucking and rail slots to minimize vessel idle time and cargo dwell durations. From the perspective of carriers, the precision of ETA predictions is indispensable for maintaining schedule integrity, ensuring seamless transshipment connectivity, and informing strategic voyage speed adjustments \citep{yan2021emerging}. Conversely, the propagation of ETA discrepancies, which is often exacerbated by port congestion or unforeseen systemic disruptions, results in significant operational inefficiencies characterized by either underutilized berths or excessive vessel queuing at anchorage. The operational inefficiencies lead to resource wastage and degraded transportation throughput \citep{chu2024evaluation, yang2024efficient}.

\textcolor{black}{Generally, existing studies in maritime logistics have largely concentrated on the next-port ETA prediction task, which estimates the remaining sailing time of the current voyage segment. Supported by AIS data, vessel schedules, and operational information, both deep learning and classical machine learning methods have achieved substantial improvements in this task. In contrast, future-port ETA prediction has received comparatively limited attention. This task is nevertheless of considerable practical relevance to supply chain reliability. In liner shipping, deteriorating schedule reliability has generated significant costs for carriers and shippers, who require more accurate advance ETAs for future ports to support slot allocation, empty container allocation, and transshipment arrangements \citep{karmelic2025liner, okur2022schedule}. Concretely, from the carrier's perspective, reliable ETAs at multiple downstream ports can support vessel operation planning, transshipment coordination, feeder connections, and service reliability management. From the shipper's perspective, early and credible ETA information enables proactive delay anticipation, procurement adjustment, and contingency planning, thereby reducing inventory risks and improving supply chain continuity. However, these operational needs cannot be adequately addressed by next-port ETA prediction alone, since cargo delivery often involves multiple subsequent voyage segments. Therefore, forecasting the sailing durations of future voyage segments is a necessary step toward providing reliable downstream arrival expectations, strengthening service commitments and enhancing supply chain resilience.}

\textcolor{black}{Compared with next-port ETA prediction, predicting sailing durations for future voyage segments is substantially more challenging. Firstly, as the prediction horizon extends, the vessel's current AIS-reported state contributes marginally to predicting the vessel's future navigation status. Although some studies attempt to infer sailing times by predicting short-term vessel trajectories or speeds from the current state and then combining them with geographical distances \citep{8294051, CHU2025105128, WANG2025121873}, such approaches are primarily designed for real-time or near-real-time ETA prediction over short horizons, often within only a few days, and therefore difficult to generalize to the prediction of future voyage segments several days or even weeks ahead. Secondly, the vessel's AIS-reported ETA has been demonstrated as one of the most significant features in next-port ETA prediction \citep{CHU2025105128}. For future voyage segments, however, this key feature is inherently unavailable, requiring the model to infer sailing durations without one of the most direct real-time indicators used in next-port ETA prediction. Thirdly, the extended time horizon introduces much stronger uncertainty into the prediction task such as future port congestion levels, which makes the sailing time of future segments considerably harder to estimate accurately.}


\textcolor{black}{To address these gaps, we develop a segment-wise forecasting framework for the sailing duration prediction task on future voyage segments. First, to mitigate the limited utility of the vessel's current navigation state for long-horizon prediction, we shift the analytical perspective from vessel-wise prediction to segment-wise prediction and explicitly model the historical sailing duration sequences of voyage segments, which contain temporal regularities informative for downstream forecasting. Second, because AIS-reported ETA indicators are unavailable for future ports, we capture temporal patterns from historical segment-duration sequences and port congestion evolution, and use these time-series features to substitute for part of the missing predictive information. Third, to account for the increasing uncertainty induced by the extended forecasting time horizon, we formulate a joint learning problem that simultaneously predicts future segment sailing durations and future port congestion states. By exploiting the shared latent dynamics between sailing-duration evolution and port operating conditions, the proposed multi-task framework yields more robust long-horizon segment-level sailing duration forecasts and provides stronger support for maritime planning. The specific contributions of this paper are summarized as follows:}

\begin{enumerate}
    \item We focus on a new prediction task targeting the sailing durations of future voyage segments on liner services, \textcolor{black}{which departs from the conventional focus on} estimating only the remaining time to the next port. This formulation directly supports reliability-oriented planning \textcolor{black}{by enabling downstream ETA predictions across multiple future ports, thereby facilitating carrier-side operational coordination and shipper-side proactive supply-chain adjustment}.
    \item We develop a unified sequence-to-sequence (Seq2Seq) transformer-based architecture with a multi-task learning strategy for segment-level sailing duration predictions. To address the lack of real-time AIS data on future legs, we transform the view from vessel-level to segment-level to avoid the dependency on AIS data. Key features including segment-level historical sailing duration sequences, port congestion levels, and segment identifiers are selected to model each voyage segment. The masked attention mechanism is utilized to capture the long-range time dependency of sailing durations. By jointly predicting sailing durations and destination port congestion levels with multi-task learning strategy, the model captures shared latent signals that govern both sailing behavior and port operating conditions under limited observations.
    \item We evaluate the proposed framework on a real-world global sailing record dataset in 2021. Extensive experiments demonstrate that the proposed approach achieves consistently strong predictive accuracy. It outperforms multiple competitive baselines, including gradient boosting models and sequential deep learning models, across segments with various sailing distances and service frequencies.
\end{enumerate}

\begin{figure}[htbp]
    \centering
    \includegraphics[width=0.95\textwidth]{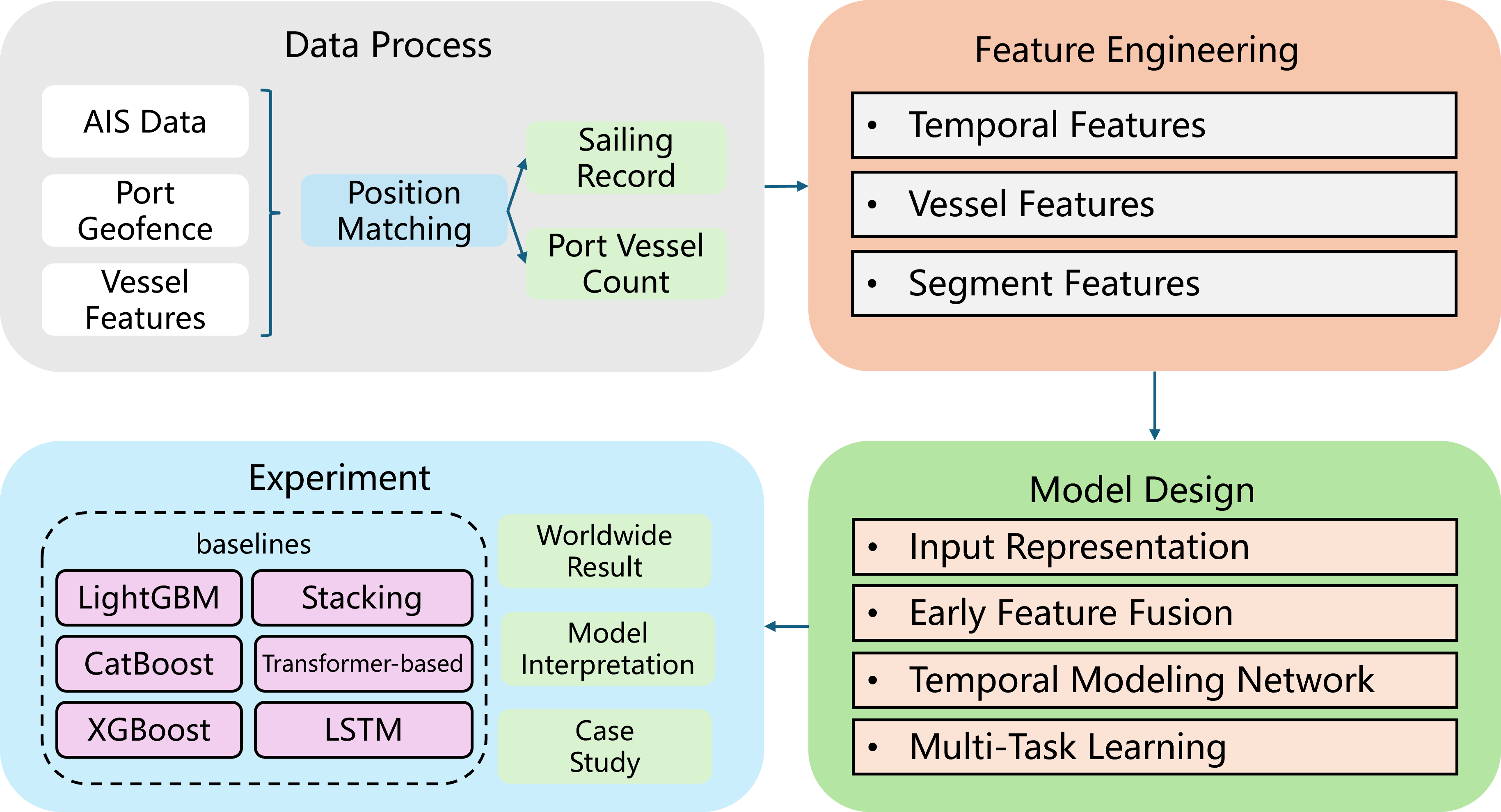}
    \caption{\textcolor{black}{An overview of the proposed framework.}}
    \label{fig:framework}
\end{figure}

Figure~\ref{fig:framework} presents an overview of the proposed framework. The remainder of this paper is organized as follows. Section~\ref{sec:literature} reviews the related literature and identifies the research gaps addressed in this study. Section~\ref{sec:method} describes the data preprocessing pipeline, the feature engineering, and the proposed model architecture. Section~\ref{sec:experiment} reports the experimental setup and validation results on voyage segments worldwide. Finally, Section~\ref{sec:conclusion} concludes the paper with a discussion of limitations and directions for future research.

\section{Literature Review}
\label{sec:literature}
 
This study belongs to the ETA prediction problem in maritime transportation. We first review the maritime ETA literature to summarize the main methodological approaches. Given the structural similarities between inland ETA prediction and the maritime setting, we then survey representative inland approaches to extract transferable modeling insights. Finally we clarify the research gap.

\subsection{Maritime ETA Prediction}

Research on maritime ETA predictions can be \textcolor{black}{broadly} divided into two categories, path-finding algorithms and data engineering algorithms \citep{noman2025review}. The former \textcolor{black}{aimed} to predict the remaining trajectories from the current position to the destination port, afterwards calculating the remaining sailing time based on the predicted vessel trajectory and the vessel speed \citep{8294051, kwun2021prediction, PARK2021100012}. Such approaches \textcolor{black}{described} the future behavior of vessels in detail but may suffer from the sparsity of trajectories. In contrast, the latter \textcolor{black}{tried} to predict the result directly without explicitly reconstructing the voyage path \citep{chu2024evaluation, el2022machine, kolley2023robust, servos2019travel, 10.1109/ITSC48978.2021.9564883}. These methods \textcolor{black}{extracted} meaningful features from real-time AIS data such as positions and speeds, combined with vessels' static features, to learn nonlinear relationships between features and remaining sailing durations.

\citet{10.1109/ITSC48978.2021.9564883} \textcolor{black}{benchmarked} three learning paradigms including gradient boosting decision trees (GBDT), multi-layer perceptron (MLP), and gated recurrent unit (GRU) for ETA predictions on inland natural and artificial waterways, reporting that the GRU achieved the best accuracy with the root mean square error (RMSE) of 8.50 minutes and mean absolute error (MAE) of 5.92 minutes. \citet{ABDI2024123988} \textcolor{black}{proposed} a fusion based deep learning framework for vessel arrival time predictions that \textcolor{black}{integrated} multiple data sources including AIS trajectories, vessel attributes, maritime weather, and ocean currents into a unified model. The architecture first \textcolor{black}{extracted} local temporal patterns from each time dependent stream with convolutional neural networks (CNN), then \textcolor{black}{captured} long range dependencies via long short-term memory (LSTM) augmented with attentions, and finally \textcolor{black}{concatenated} static vessel features before dense layers. The result \textcolor{black}{showed} the RMSE of 10.63 and mean absolute percentage error (MAPE) of 35.11\%. \citet{app11104410} \textcolor{black}{introduced} a two-stage ETA framework that explicitly \textcolor{black}{incorporated} future weather information. In the first stage, the route \textcolor{black}{was} inferred by selecting historical trajectories whose observed weather patterns most closely \textcolor{black}{matched} the forecasting future conditions. In the second stage, the voyage speed \textcolor{black}{was} estimated using a Bayesian learning approach. 28\% improvement in arrival time prediction accuracy \textcolor{black}{was} reported compared with previous methods. \citet{CHU2025105128} \textcolor{black}{proposed} a stacking model framework for ETA prediction that \textcolor{black}{fused} static port call records with dynamic AIS trajectories to reduce the bias of historical actual time of arrival (ATA) and thus to enhance the ETA prediction accuracy. With the case analysis of Hong Kong, MAE \textcolor{black}{was} reduced by 54.53\% and RMSE by 50.14\% compared to the vessel-reported ETA. \citet{10.1007/978-3-031-43612-3_13} \textcolor{black}{designed} a three-layer neural network trained on historical AIS trajectories for inland waterways in Netherlands and Germany, and \textcolor{black}{reported} accuracy gains of 20.6\% for short voyage, 4.8\% for medium voyage, and 13.4\% for long-haul voyage. \citet{8294051} \textcolor{black}{proposed} a data-driven method for estimating vessels' times of arrival using historical AIS trajectories. It \textcolor{black}{built} density and direction maps and \textcolor{black}{applied} an optimized path-finding algorithm with a statistical speed model to infer routes and travel times from current vessel positions to target ports. Validated on real operational scenarios in Europe, their approach \textcolor{black}{demonstrated} clear gains over schedule-based heuristics. \citet{10422495} \textcolor{black}{addressed} ETA prediction to pilotage area with heterogeneous sources of data. The temporal convolutional network \textcolor{black}{was} selected to extract temporal patterns from sequential AIS data. Experiments on two real-world Singapore datasets yielded MAE less than 4.86 minutes and around 90\% of absolute errors within 10 minutes. \citet{PARK2021100012} \textcolor{black}{proposed} an AIS-driven ETA system with two stages. In the first stage, possible vessel trajectories \textcolor{black}{were} learned through a path-finding module enhanced with reinforcement learning. In the second stage, speed-over-ground \textcolor{black}{was} estimated using a Markov-chain formulation with Bayesian sampling. Experiments on the Busan port dataset \textcolor{black}{demonstrated} the advantage. \citet{bulk} \textcolor{black}{studied} ETA predictions for bulk ports by framing the problem as sequence learning over historical AIS data. Multiple deep learning models \textcolor{black}{were} trained and compared. Experiments \textcolor{black}{showed} that these sequential models \textcolor{black}{could} effectively extract features from vessels' historical trajectories and enhance the prediction accuracy of remaining sailing time to the destination port.

There are two limitations in the above studies. The first and most significant limitation is that the prediction model depends \textcolor{black}{heavily} on real-time AIS data of current voyage. The accuracy of predictions degenerates significantly if the AIS data is unavailable \citep{CHU2025105128}. It determines that the models above can only be used for the sailing duration prediction task on the current segment but not future voyage segments. Besides, the studies above do not make full use of the sailing duration time series on each segment. For each voyage segment, historical sequences of sailing durations involve changing patterns which are useful for long-term forecasting. Explicitly modeling the sequence of sailing durations helps the model to extract inherent features and enhance ETA prediction accuracy.

\subsection{Inland ETA Prediction}

Inland transportation has some features in common with maritime transportation. Both domains rely on trajectory data sources. In maritime transportation, vessel movements are captured by AIS data, whereas in inland logistics, vehicle trajectories are collected through GPS devices. Similarly, maritime voyage segments between ports correspond to road segments in land traffic networks. ETA in both situations is influenced by congestion factors. In the field of road transportation, researchers have extensively applied temporal forecasting models to capture traffic dynamics and to predict the arrival time of vehicles. Given these similarities, many time-series prediction techniques developed for inland ETA prediction tasks can provide methodological insights for maritime applications.

\citet{liu2024itransformerinvertedtransformerseffective} \textcolor{black}{introduced} iTransformer, an inverted attention architecture that \textcolor{black}{treated} variables as tokens, enabling the model to learn cross variable dependencies effectively. This design \textcolor{black}{was} able to capture relationships in traffic-flow dynamics on approaching routes, reporting to deliver strong long-term forecasts on standard benchmarks that \textcolor{black}{included} traffic data. \citet{Derrow_Pinion_2021} \textcolor{black}{presented} a graph neural network (GNN) framework for ETA prediction that \textcolor{black}{was} deployed in Google Maps, yielding substantial accuracy gains over prior production models. The approach \textcolor{black}{partitioned} the road network into spatially adjacent supersegments and \textcolor{black}{constructed} a directed spatiotemporal graph in which nodes \textcolor{black}{represented} supersegments and edges \textcolor{black}{encoded} traffic influence. By learning interactions across neighboring supersegments and capturing the network-wide congestion propagation, the method \textcolor{black}{improved} robustness under dynamic traffic conditions. \citet{liu2023uncertainty} \textcolor{black}{addressed} the critical need for quantifying prediction reliability in ride-hailing services by proposing an uncertainty-aware probabilistic deep learning model called ProbTTE for travel time predictions. The framework not only \textcolor{black}{generated} accurate point estimations but also \textcolor{black}{yielded} the parameters of a probability distribution, effectively capturing the inherent variability and uncertainty in urban traffic.
 
However, there are some gaps between inland transportation and maritime transportation. Firstly, road vehicles must follow predefined links in a road network, so the travel time is largely determined by paths constrained to the road graph. By contrast, maritime voyages are weakly constrained by fixed trajectories and vessels can select among multiple routes. As a result, the sailing time between two ports can vary substantially due to the route choice and the differences in nautical distance. Secondly, congestion in inland systems typically \textcolor{black}{happens} on the road. In maritime settings, congestion concentrates at ports but not along the voyage segments. Vessels generally adjust speed when approaching the destination port, responding to berth availability and port operations. Thirdly, sailing times are significantly influenced by vessel-specific attributes. This leads to high variance in sailing durations even for vessels departing on the same voyage segment at proximate times, a phenomenon markedly different from the relatively stable travel times observed on road segments.

\subsection{Research Gap}

In summary, while contemporary maritime ETA prediction literature has achieved significant accuracy, the majority of studies remain confined to estimating the remaining sailing duration of the current segment by leveraging real-time AIS data and immediate vessel states. The heavy reliance on real-time AIS information inherently restricts the predictive horizon and compromises model robustness in scenarios where signals are delayed, absent, or unavailable for future segments. Furthermore, existing methodologies often overlook the intrinsic temporal regularities embedded within historical segment-level sailing duration sequences, which are critical assets for effective long-term forecasting. Given the extended temporal span of future voyage segments, immediate vessel-specific behavior features provide diminishing marginal utilities.

To bridge these gaps, this study investigates the task of predicting sailing durations for future voyage segments, a domain that has yet to be explored in depth. We pivot the analytical focus from real-time vessel tracking to modeling the long-range temporal dependencies inherent in segment-level sailing duration time series. Specifically, we propose a transformer-based model designed to extract latent evolutionary patterns from historical sailing records, which are then integrated with port congestion proxies, static vessel characteristics and segment identifiers to deliver reliable forecasts across global liner services. To further enhance the predictive stability, a multi-task learning strategy is adopted to jointly predict destination port congestion and sailing durations on the voyage segments, thereby capturing the correlated latent signals that govern both sailing behavior and port operational efficiency.

\section{Method}
\label{sec:method}
\subsection{Problem Definition}
\label{sec:problem}

In contrast to conventional vessel-level next-port ETA predictions, we reformulate the task into a time-series forecasting problem at the voyage segment level. Table~\ref{tab:segment_view} shows the segment-level data format of sailing records. Specifically, continuous time is discretized into non-overlapping, fixed-length time windows of duration $\Delta t$ with integer indices $t \in \mathbb{Z}_{>0}$. Time window $t$ defines the half-open interval $[(t-1)\Delta t,\ t\Delta t)$ which guarantees each timestamp belongs to exactly one time window and adjacent time windows share no overlap. Let $\mathcal{P} = \{p^{(1)}, p^{(2)}, \dots, p^{(N)}\}$ denote the set of ports where $N$ is the number of ports. $p^{(i)}$ represents the \textit{i}th port. Let $\mathbf{A} = [a^{(ij)}]$ stand for the adjacency matrix.

\begin{equation*}
\mathbf{A}(i,j)=\begin{cases}
1,& \text{if segment from}\ p^{(i)}\ \text{to}\ p^{(j)}\ \text{exists}\\
0,& \text{otherwise},
\end{cases}
\qquad \mathbf{A}\in\{0,1\}^{N\times N}
\end{equation*}

\begin{table}[h]
\centering
\caption{An example of sailing record data format on the segment-level view}
\label{tab:segment_view}
\begin{tabularx}{\textwidth}{XXll}
\toprule
\textbf{Segment} & \textbf{Start\ time} & \textbf{Vessel\ ID} & \textbf{Sailing\ time\ (h)} \\
\midrule
Hong Kong $\rightarrow$ Singapore & 2021-01-10 06:00:00 & 10254 & 120.00 \\
Hong Kong $\rightarrow$ Singapore & 2021-01-11 12:00:00 & 10069 & 105.00 \\
Hong Kong $\rightarrow$ Singapore & 2021-01-12 18:00:00 & 10236 & 95.00 \\
\bottomrule
\end{tabularx}
\end{table}

Let $\mathcal{E} = \{e^{(ij)}\}$ denote the set of voyage segments in a maritime liner network, where each segment $e^{(ij)} = (\textcolor{black}{p^{(i)}} \rightarrow \textcolor{black}{p^{(j)}})$ connects two consecutive ports $\textcolor{black}{p^{(i)}}$ to $\textcolor{black}{p^{(j)}}$ and $|\mathcal{E}|=K$, where $K$ is the total number of voyage segments. For each segment $e^{(ij)}$, historical sequences of sailing durations are collected as $\mathbf{Y}_{T-L:T}^{(ij)} = [ \mathbf{Y}_{T-L}^{(ij)}, \mathbf{Y}_{T-(L-1)}^{(ij)}, \dots, \mathbf{Y}_{T-1}^{(ij)} ]$ indexed by the departure time window $t$, where $T$ denotes the time window index of timestamp to make predictions separating the lookback period and forecasting horizon, $L$ denotes the length of the lookback period, and $\mathbf{Y}_t^{(ij)}$ represents the sailing duration on segment $e^{(ij)}$ with the departure in time window $t$. In addition, both time-varying covariates and static features are incorporated to enhance the forecasting accuracy. Some covariates can only be observed up to the time window $T$, such as historical port congestion levels. These covariates are denoted as $\mathbf{X}_{T-L:T}^{(ij)} = [ \mathbf{X}_{T-L}^{(ij)}, \mathbf{X}_{T-(L-1)}^{(ij)}, \dots, \mathbf{X}_{T-1}^{(ij)} ]$, where $\mathbf{X}_t^{(ij)}$ represents the observable covariates on segment $e^{(ij)}$ in time window $t$ before time window $T$. Other features can be known for both past and future periods, for instance, the weekday of the departure time window and static vessel characteristics. These variables are denoted as $\mathbf{S}_{T-L:T+H}^{(ij)} = [ \mathbf{S}_{T-L}^{(ij)}, \mathbf{S}_{T-(L-1)}^{(ij)}, \dots, \mathbf{S}_{T}^{(ij)}, \mathbf{S}_{T+1}^{(ij)}, \dots, \mathbf{S}_{T+(H-1)}^{(ij)} ]$ where $H$ represents the length of forecasting horizon. $\mathbf{S}_t^{(ij)}$ represents these feature variables on segment $e^{(ij)}$ in time window $t$. By jointly modeling $\mathbf{Y}_{T-L:T}^{(ij)}$, $\mathbf{X}_{T-L:T}^{(ij)}$ and $\mathbf{S}_{T-L:T+H}^{(ij)}$, the predictive framework can leverage historical dynamic information while conditioning on known future static features, therefore improving the long-term prediction accuracy and stability.

The objective of the framework is to predict the sailing durations $\mathbf{Y}_{T:T+H}^{(ij)}$ associated with the corresponding covariates $\mathbf{X}_{T:T+H}^{(ij)}$ for future horizons $H$,
\[
\hat{\mathbf{Y}}_{T:T+H}^{(ij)} = [\hat{\mathbf{Y}}_{T}^{(ij)}, \hat{\mathbf{Y}}_{T+1}^{(ij)}, \dots, \hat{\mathbf{Y}}_{T+(H-1)}^{(ij)}], 
\]
\[
\hat{\mathbf{X}}_{T:T+H}^{(ij)} = [\hat{\mathbf{X}}_{T}^{(ij)}, \hat{\mathbf{X}}_{T+1}^{(ij)}, \dots, \hat{\mathbf{X}}_{T+(H-1)}^{(ij)}], 
\]
where $\hat{\mathbf{Y}}_{t}^{(ij)}$ is the predicted sailing duration of segment $e^{(ij)}$ at future time window $t$ and $\hat{\mathbf{X}}_{t}^{(ij)}$ is the predicted \textcolor{black}{value} for corresponding covariates on segment $e^{(ij)}$ at future time window $t$. The model thus learns a mapping
\[
f: \big(\mathbf{Y}_{T-L:T}^{(ij)}, \mathbf{X}_{T-L:T}^{(ij)}, \mathbf{S}_{T-L:T+H}^{(ij)}\big) \mapsto \big(\hat{\mathbf{Y}}_{T:T+H}^{(ij)}, \hat{\mathbf{X}}_{T:T+H}^{(ij)}\big),\ \text{for}\ e^{(ij)} \in \mathcal{E}
\]

\begin{figure}[htbp]
    \centering
    \includegraphics[width=0.95\textwidth]{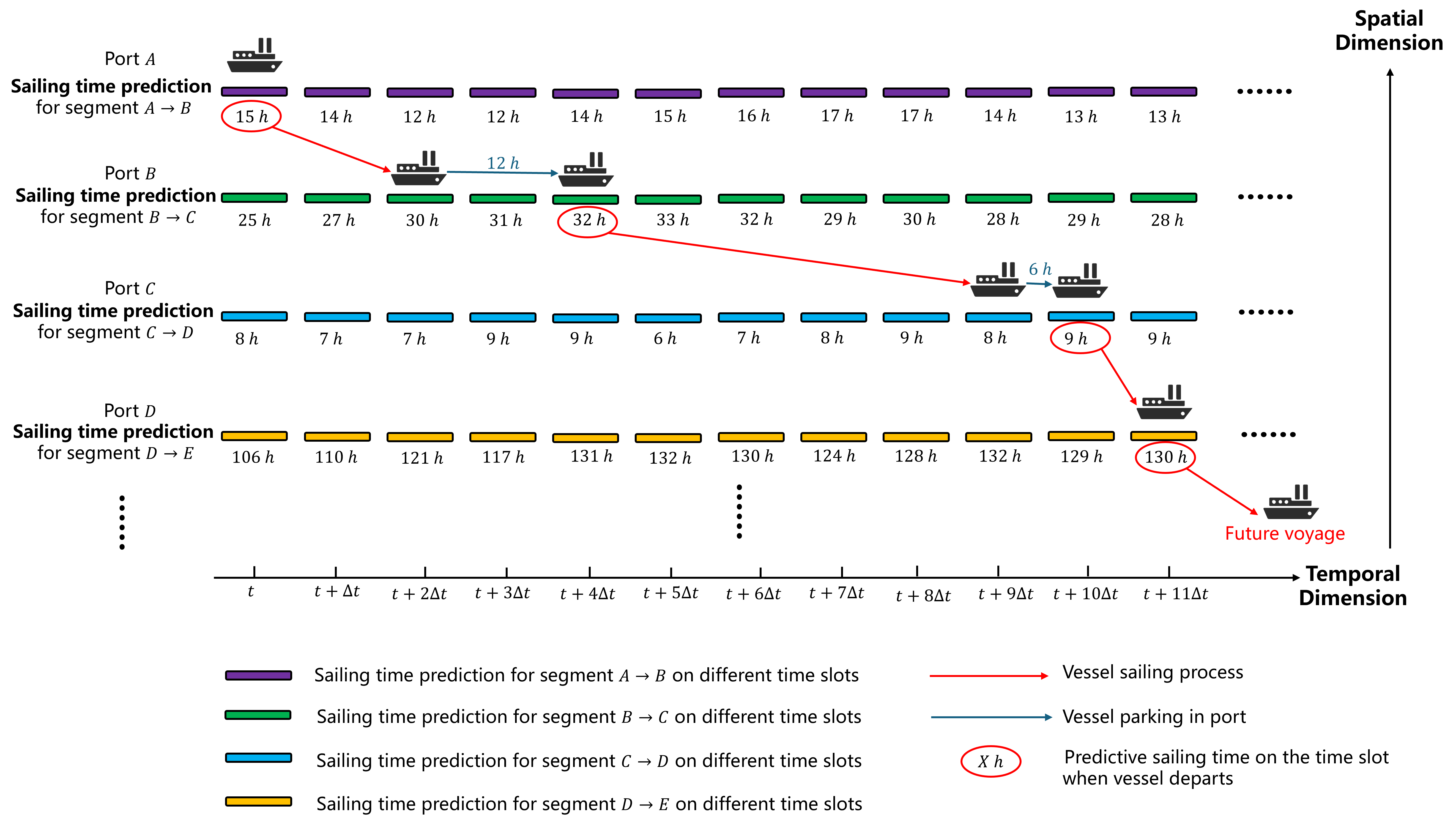}
    \caption{\textcolor{black}{Illustration of multi-segment ETA estimation via spatiotemporal matching ($\Delta t = 6$ hours).}}
    \label{fig:temporal_spatial_search}
\end{figure}

This formulation allows the model to provide segment-level sailing time forecasting over the future horizons. Once the predicted sailing time is obtained for a given segment $e^{(ij)}$ over the horizon from time window $T$ to $T+(H-1)$, the estimated voyage time of a particular vessel on segment $e^{(ij)}$ can be derived by aligning the departure timestamp with the corresponding time window. In this way, the framework changes the view from vessel-level to segment-level, enabling one model to be used for different vessels, achieving more robust and stable predictions. \textcolor{black}{Furthermore, cumulative error can also be considered under the framework. As described in Figure~\ref{fig:temporal_spatial_search}, for a given vessel, the sailing time on a segment is obtained by matching its departure timestamp to the corresponding future time window. Accordingly, the predicted sailing time of one segment can be used to update the estimated departure time for the next segment, allowing upstream delays to propagate by shifting the matched future time state of downstream segments.}

\subsection{Data Preprocessing}
\label{sec:process}

To construct such a model, sailing durations of each voyage segment and port congestion levels are essential features. Therefore, it is necessary to first establish a data preprocessing pipeline that transforms raw data into the structured voyage record dataset and port vessel count dataset, serving for further utilization. The raw data contains AIS data, port geofence data and static vessel information. The basic items of raw data above are described in Appendix~\ref{sec:raw}.

\begin{figure}[h]
    \centering
    \includegraphics[width=0.95\textwidth]{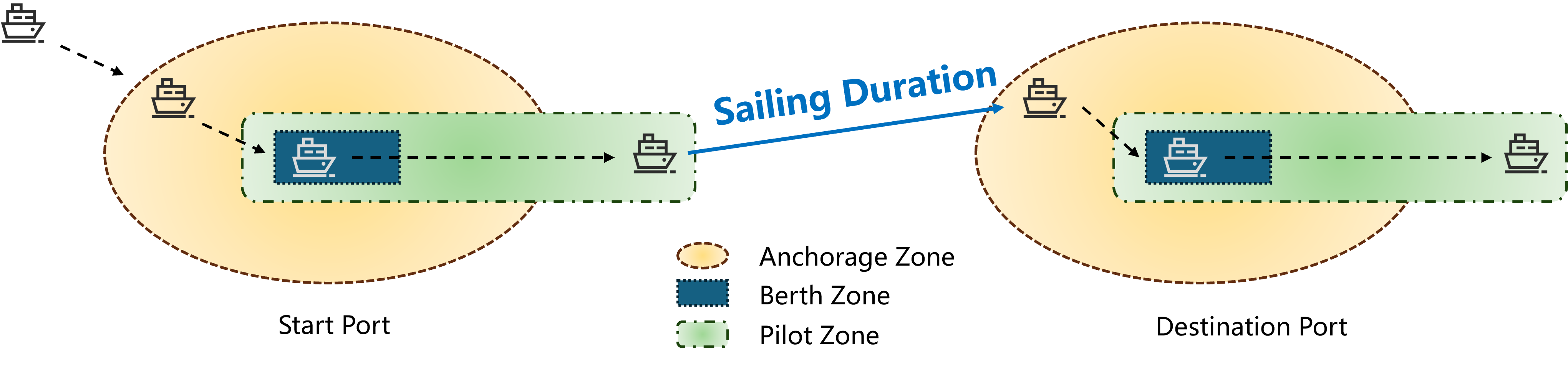}
    \caption{The schematic diagram of vessel departure and arrival process.}
    \label{fig:port_process}
\end{figure}

It is important to give an accurate and objective definition \textcolor{black}{of} the sailing duration because it serves as the target label for model predictions. As illustrated in Figure~\ref{fig:port_process}, the general process of vessel departure and arrival involves several major spatial zones, including the departure \textcolor{black}{port's} pilotage area, open sea area, destination \textcolor{black}{port's} anchorage area, destination \textcolor{black}{port's} pilotage area and destination \textcolor{black}{port's} berth. Since the sailing duration is intended to characterize vessels' navigation behavior in open sea area rather than within near-port waters, the sailing duration is defined as the time elapsed between leaving the departure \textcolor{black}{port's} pilotage area and arriving at the destination \textcolor{black}{port's} anchorage area.

Based on this definition, the port geofence data can be used to segment vessel AIS trajectories into separate voyages, allowing each trajectory segment to be aggregated into a structured voyage record. Specifically, for each vessel's complete AIS data, the vessel's latitude-longitude coordinates are first intersected with the berth boundaries contained in the port geofence dataset to identify the departure and destination ports for each voyage. For any two consecutive berthing events, the AIS trajectories are further intersected with the pilotage boundary of the departure port and the anchorage boundary of the destination port. The last AIS point within the departure port's pilotage area is regarded as the departure mark, and the first AIS point within the destination port's anchorage area is regarded as the arrival mark. The time difference between these two points is recorded as the sailing duration of the voyage. \textcolor{black}{Due to factors such as incorrectly identified port-call events or unexpected incidents during navigation, some voyage records exhibit sailing durations that are substantially longer than the typical level of the corresponding segment. To mitigate the influence of such extreme observations, segment-wise outlier filtering is performed by removing voyage records whose sailing durations exceed the $97.5th$ percentile of the sailing time distribution for the corresponding segment.} Finally, by matching the vessel's International Maritime Organization (IMO) identifier with the static vessel information, the corresponding vessel characteristics can be attached to each voyage record. Subsequently, voyage records are synchronized onto the discrete timeline by aligning each voyage's actual departure timestamp to its unique time window $t = [(t-1)\Delta t,\ t\Delta t)$. In this way, a regularized voyage record dataset is constructed.

On this basis, \textcolor{black}{the construction of} the port vessel count dataset becomes relatively straightforward. Specifically, assuming that the initial number of vessels in each port is zero, the voyage record dataset which records voyage segments, vessel departure time and arrival time can be used to calculate the number of vessels arriving at and departing from each port within every time window $t$, therefore deriving the relative vessel count for each port over time.

Through this pipeline, structured datasets of regularized voyage records and port vessel counts can be obtained for subsequent analysis and modeling.

\subsection{Feature Engineering}

Selecting appropriate and informative features is crucial for improving the predictive accuracy of the model. For the task of long-term sailing duration predictions on future voyage segments, three categories of features are primarily considered: temporal features, vessel features and segment features. Table~\ref{tab:feature_summary} summarizes the selected input features. The details for each feature group are described in the following subsections.

\begin{table}[htbp]
\centering
\caption{Summary of input features for voyage segment duration prediction.}
\label{tab:feature_summary}
\renewcommand{\arraystretch}{1.2}
\begin{tabularx}{\textwidth}{XlXl}
\toprule
\textbf{Feature Name} & \textbf{Category} & \textbf{Notation} & \textbf{Time Span} \\
\midrule
Historical sailing duration sequence & Temporal & $\mathbf{Y}_t^{(ij)} = \{y_t^{(ij)}\}$ & $t \in [T-L, T-1]$ \\
Historical vessel count sequence at destination port & Temporal & $\mathbf{X}_t^{(ij)} = \{x_t^{(j)}\}$ & $t \in [T-L, T-1]$ \\
Time window identifier & Temporal & $\mathbf{I}_t = \{g_t, r_t\}$ & $t \in [T-L, T+(H-1)]$ \\
Vessel static characteristics & Vessel & $\mathbf{V}_t^{(ij)} = \{l_t^{(ij)}, w_t^{(ij)}, u_t^{(ij)}, c_t^{(ij)}\}$ & $t \in [T-L, T+(H-1)]$ \\
Segment identifier & Segment & $\mathbf{R}_t^{(ij)} = \{p^{(i)}, p^{(j)}, m_t^{(ij)}\}$ & $t \in [T-L, T+(H-1)]$ \\
\bottomrule
\end{tabularx}
\end{table}

\subsubsection{Temporal Features}

Temporal features are designed to capture the dynamic changing \textcolor{black}{patterns} of both segment-level sailing durations and port congestion over time. In the study, two major time dependent sequential variables are employed. The first is the historical time series of sailing durations for each voyage segment. This variable reflects the temporal continuity and periodic variation of sailing efficiency, which may be influenced by seasonal patterns. By modeling the sequential correlation among past sailing duration sequences, the model can learn long-term temporal dependencies and better predict future voyage durations. The feature of sailing durations for each voyage segment can be denoted as $y_t^{(ij)}$, representing the sailing duration of segment $\textcolor{black}{p^{(i)}} \rightarrow \textcolor{black}{p^{(j)}}$ with departure timestamp in time window $t$. The sequence $[y_{T-L}^{(ij)}, y_{T-(L-1)}^{(ij)}, \dots, y_{T-1}^{(ij)}]$ constructs the sequential feature $\mathbf{Y}_{T-L:T}^{(ij)}$ defined in Section~\ref{sec:problem}. \textcolor{black}{Notably, since the observed sailing durations are realized outcomes under prevailing navigation conditions, the sequence can also implicitly capture part of the regular meteorological effects.}

The second sequential temporal feature is the historical time series of the vessel count at the destination port. This feature serves as a proxy for port congestion, representing the operational load of the destination port at each time window. The congestion level directly affects the decision of the captain. Captains tend to lower down the speed of the vessel to save the energy if they are informed the congestion of the destination port because they have to wait in queue to get the service. Combining the time series allows the model to capture not only seasonal patterns within sailing duration sequences but also external factors caused by port operations. Together, these temporal features enable the model to effectively integrate sequential information and external temporal dependencies for more accurate long-term sailing duration forecasting. The feature of vessel counts at the destination port can be denoted as $x_t^{(j)}$, representing the vessel count at the destination port $j$ in time window $t$. The sequence $[x_{T-L}^{(j)}, x_{T-(L-1)}^{(j)}, \dots, x_{T-1}^{(j)}]$ constructs the sequential feature $\mathbf{X}_{T-L:T}^{(ij)}$ defined in Section~\ref{sec:problem}. $x_t^{(j)}$ is not related to the start port of the segment $\textcolor{black}{p^{(i)}}$, so it is indexed with the destination port $\textcolor{black}{p^{(j)}}$ only. \textcolor{black}{To further validate the appropriateness of the port congestion proxy, we provide comparisons with alternative congestion proxies in Appendix~\ref{sec:congestion_indicator}.}

Beyond the two aforementioned sequential temporal features, the weekday of the departure time window $g_t$ and its positional order index within the day $r_t$ are selected as identifiers for each time window $t$. Let $\mathbf{I}_{t} = \{g_t, r_t\}$ denote the identifier of time window $t$. The sequence $\mathbf{I}_{T-L:T+H} = [\mathbf{I}_{T-L}, \mathbf{I}_{T-(L-1)}, \dots, \mathbf{I}_{T}, \mathbf{I}_{T+1}, \dots, \mathbf{I}_{T+(H-1)}]$ belongs to the feature $\mathbf{S}_{T-L:T+H}^{(ij)}$ defined in Section~\ref{sec:problem}. $\mathbf{I}_{t}$ only depends on the time window $t$ so that the index for segments is omitted.

\subsubsection{Vessel Features}

Vessel features describe the physical characteristics of \textcolor{black}{vessels} that may influence their sailing durations. These static attributes remain constant for a given vessel and serve as important features for sailing duration predictions. Key features such as the vessel length, vessel width and vessel twenty-foot equivalent unit (TEU) are selected. These features describe the size and capacity of vessels, indicating different voyage patterns. Besides, the carrier of the vessel serves as a significant feature. Different carriers prefer different routes even for the same voyage segment. Therefore, the different choices of routes cause that the actual sailing distance becomes different among carriers, resulting in the variance of sailing durations. Vessel features including vessel length, vessel width, vessel TEU, and vessel carriers are denoted as $\mathbf{V}_{t}^{(ij)}=\{l_t^{(ij)}, w_t^{(ij)}, u_t^{(ij)}, c_t^{(ij)}\}$, representing the vessel's features $\mathbf{V}_{t}^{(ij)}$ including the vessel length $l_t^{(ij)}$, vessel width $w_t^{(ij)}$, vessel TEU $u_t^{(ij)}$, and vessel carriers $c_t^{(ij)}$ on segment $\textcolor{black}{p^{(i)}} \rightarrow \textcolor{black}{p^{(j)}}$ with the departure timestamp in time window $t$. The sequence $\mathbf{V}_{T-L:T+H}^{(ij)} = [\mathbf{V}_{T-L}^{(ij)}, \mathbf{V}_{T-(L-1)}^{(ij)}, \dots, \mathbf{V}_{T}^{(ij)}, \mathbf{V}_{T+1}^{(ij)}, \dots, \mathbf{V}_{T+(H-1)}^{(ij)}]$ belongs to the feature $\mathbf{S}_{T-L:T+H}^{(ij)}$ defined in Section~\ref{sec:problem}.

\subsubsection{Segment Features}

Different segments have their inherent characteristics, which determine the variability of sailing durations and their sensitivity to external conditions. The changing pattern of voyage time and its correlation with port congestion may vary significantly across different segments. To capture such heterogeneity, several segment-level features are incorporated into the model. The start port and destination port serve as the unique identifier of the segment. Besides, the arriving terminal of the destination port is selected, considering different congestion levels of different terminals in the same port. Segment features including the start port, destination port and destination terminal are denoted as $\mathbf{R}_{t}^{(ij)}=\{p^{(i)}, p^{(j)}, m_t^{(ij)}\}$, where $\mathbf{R}_{t}^{(ij)}$ represents the features of segment $\textcolor{black}{p^{(i)}} \rightarrow \textcolor{black}{p^{(j)}}$ with departure timestamp in time window $t$, including the start port $p^{(i)}$, destination port $p^{(j)}$ and destination terminal $m_t^{(ij)}$. The sequence $\mathbf{R}_{T-L:T+H}^{(ij)} = [\mathbf{R}_{T-L}^{(ij)}, \mathbf{R}_{T-(L-1)}^{(ij)}, \dots, \mathbf{R}_{T}^{(ij)}, \mathbf{R}_{T+1}^{(ij)}, \dots, \mathbf{R}_{T+(H-1)}^{(ij)}]$ belongs to the feature $\mathbf{S}_{T-L:T+H}^{(ij)}$ defined in Section~\ref{sec:problem}. It is notable that features $p^{(i)}$ and $p^{(j)}$ do not change with the time window $t$ since the start port and destination port remain the same for the selected segment. The feature $m_t^{(ij)}$ is associated with segment $\textcolor{black}{p^{(i)}} \rightarrow \textcolor{black}{p^{(j)}}$ and time window $t$ because the choice of destination terminal can vary from different segments and different time windows on the same segment.

\subsection{Model Design}

\begin{figure}[h]
    \centering
    \includegraphics[width=0.85\textwidth]{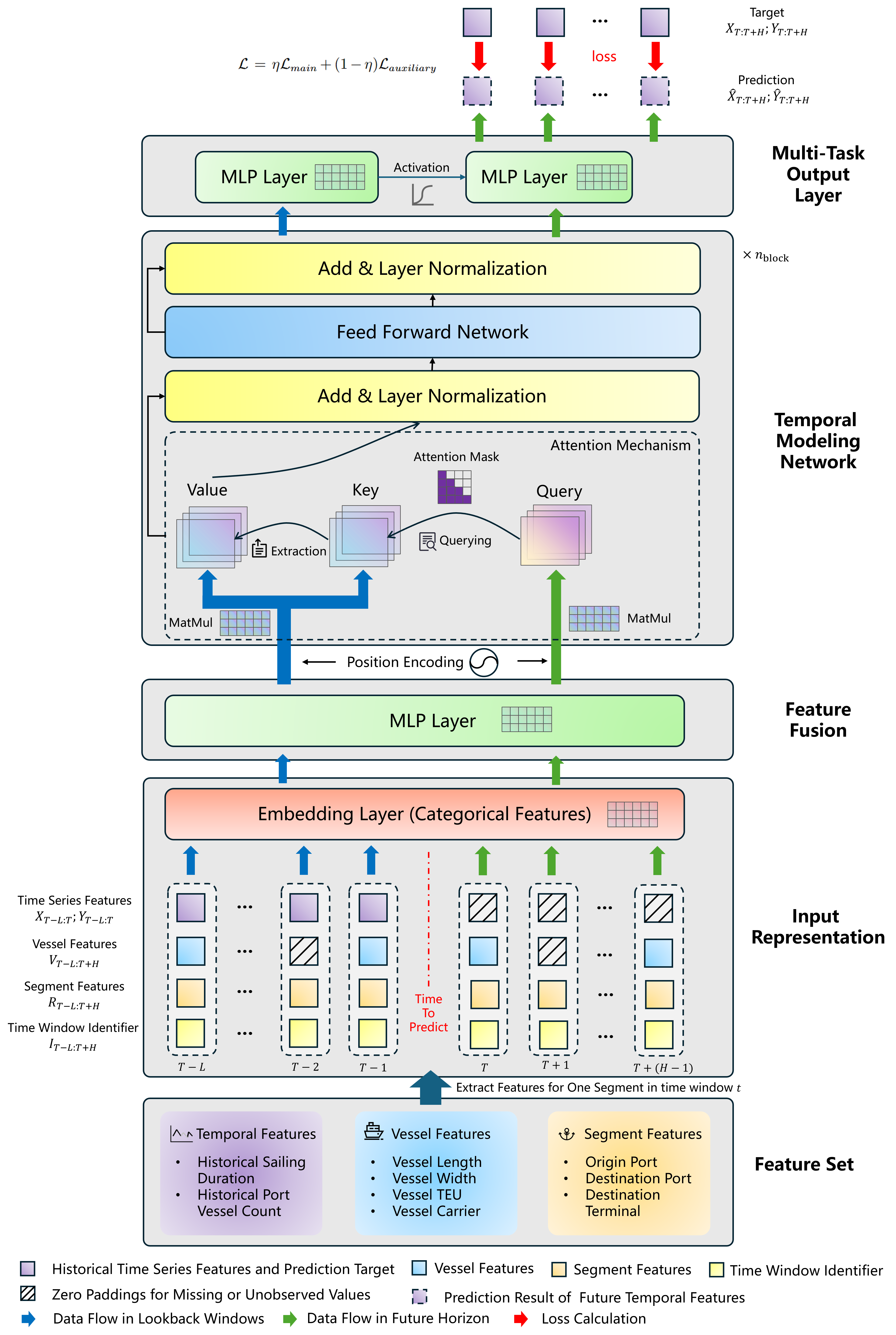}
    \caption{The overall structure of the proposed model.}
    \label{fig:model_arch}
\end{figure}

The purpose of the proposed model is to accurately predict the sailing duration of each voyage segment \textcolor{black}{over a long horizon} for future-port ETA forecasting by capturing the pattern of sailing duration time series and correlation with exogenous covariates. The overall architecture of the proposed model is shown in Figure~\ref{fig:model_arch}. Formally, the model aims to forecast the sailing duration $\mathbf{Y}_{T:T+H}^{(ij)} = [{y}_{T}^{(ij)}, {y}_{T+1}^{(ij)}, \dots, {y}_{T+(H-1)}^{(ij)}]$ on segment $\textcolor{black}{p^{(i)}} \rightarrow \textcolor{black}{p^{(j)}}$ and the vessel count $\mathbf{X}_{T:T+H}^{(ij)} = [{x}_{T}^{(j)}, {x}_{T+1}^{(j)}, \dots, {x}_{T+(H-1)}^{(j)}]$ in destination port $\textcolor{black}{p^{(j)}}$ for the future $H$ time steps with the given features $\mathbf{Y}_{T-L:T}^{(ij)}$, $\mathbf{X}_{T-L:T}^{(ij)}$ and $\mathbf{S}_{T-L:T+H}^{(ij)}=\{\mathbf{I}_{T-L:T+H}, \mathbf{V}_{T-L:T+H}^{(ij)}, \mathbf{R}_{T-L:T+H}^{(ij)}\}$.

\subsubsection{Input Representation}
\label{sec:input_representation}

For each voyage segment $\textcolor{black}{p^{(i)}} \rightarrow \textcolor{black}{p^{(j)}}$, the input features contain two types, continuous features and categorical features. For time windows without observed voyage records, the input features related to sailing records $y_t^{(ij)}$, $\mathbf{V}_{t}^{(ij)}$ and $m_t^{(ij)}$ are filled with null values. For time windows with multiple observed voyage records, one record is selected at random. Besides, due to the availability of different features, the feature length is either $L$ (e.g. $\mathbf{X}_{T-L:T}^{(ij)}$) or $L+H$ (e.g. $\mathbf{V}_{T-L:T+H}^{(ij)}$). So, it is important to introduce a module to unify the format of input features and convert the scalar features into vectors.

To address the missing data, zero is padded to the features related to sailing records in time windows with no observation. Besides, since there is no knowledge about sailing durations and vessel counts in destination ports in the future, zero padding is also added to these time series features to keep the length the same between different features:
\begin{equation*}
\tilde{\mathbf{Y}}_{T-L:T+H}^{(ij)} = \big[\, \mathbf{Y}_{T-L:T}^{(ij)} \ \| \ \mathbf{0}_{H} \,\big],
\tilde{\mathbf{X}}_{T-L:T+H}^{(ij)} = \big[\, \mathbf{X}_{T-L:T}^{(ij)} \ \| \ \mathbf{0}_{H} \,\big],
\end{equation*}
where $\mathbf{0}_{H}$ denotes an $H$-length zero vector, $\tilde{\mathbf{Y}}_{T-L:T+H}^{(ij)}$ and $\tilde{\mathbf{X}}_{T-L:T+H}^{(ij)}$ represent the corresponding sequences after padding. For categorical features $\mathbf{R}_{t}^{(ij)}=\{p^{(i)}, p^{(j)}, m_t^{(ij)}\}$, $\mathbf{I}_{t} = \{g_t, r_t\}$ and $c_t^{(ij)}$, embedding layers $\textbf{Emb}(\cdot)$ are designed for each variable to convert them from one-hot variables into dense vectors $\in\ \mathbb{R}^{d_{\mathrm{emb}}}$ with trainable parameters where $d_{\mathrm{emb}}$ is the dimension of embedding layers: 
\begin{equation*}
\tilde{\mathbf{I}}_{T-L:T+H} = \big[\, \mathrm{Concat}\big(\mathbf{Emb}_{g}(g_t), \ \ \mathbf{Emb}_{r}(r_t)\big) \,\big]_{t=T-L}^{T+(H-1)},
\end{equation*}
\begin{equation*}
\tilde{\mathbf{R}}_{T-L:T+H}^{(ij)} = \big[\, \mathrm{Concat}\big(\mathbf{Emb}_p(p^{(i)}), \ \ \mathbf{Emb}_p(p^{(j)}), \ \ \mathbf{Emb}_m(m_t^{(ij)})\big) \,\big]_{t=T-L}^{T+(H-1)},
\end{equation*}
\begin{equation*}
\tilde{\mathbf{V}}_{T-L:T+H}^{(ij)} = \big[\, \mathrm{Concat}\big(l_t^{(ij)}, w_t^{(ij)}, u_t^{(ij)}, \ \ \mathbf{Emb}_c(c_t^{(ij)})\big) \,\big]_{t=T-L}^{T+(H-1)},
\end{equation*}
\begin{equation*}
\tilde{\mathbf{S}}_{T-L:T+H}^{(ij)} = \mathrm{Concat}\big( \tilde{\mathbf{V}}_{T-L:T+H}^{(ij)}, \ \ \tilde{\mathbf{R}}_{T-L:T+H}^{(ij)}, \ \ \tilde{\mathbf{I}}_{T-L:T+H} \,\big).
\end{equation*}
This embedding scheme converts heterogeneous discrete attributes into a unified, dense representation that preserves categorical semantics while remaining compatible with downstream sequential models. $\tilde{\mathbf{I}}_{T-L:T+H}$, $\tilde{\mathbf{R}}_{T-L:T+H}^{(ij)}$, $\tilde{\mathbf{V}}_{T-L:T+H}^{(ij)}$, $\tilde{\mathbf{S}}_{T-L:T+H}^{(ij)}$ represent the corresponding feature vectors after embedding and the $\mathrm{Concat}(\cdot)$ operation combines the feature vectors at each corresponding time window across the different input sequences. To form the final input, the embedding matrix of static features $\tilde{\mathbf{S}}_{T-L:T+H}^{(ij)}$ is concatenated with time series features $\tilde{\mathbf{Y}}_{T-L:T+H}^{(ij)}$ and $\tilde{\mathbf{X}}_{T-L:T+H}^{(ij)}$:
\begin{equation*}
\mathbf{\xi}_{T-L:T+H}^{(ij)} = \mathrm{Concat}\big( \tilde{\mathbf{Y}}_{T-L:T+H}^{(ij)}, \ \ \tilde{\mathbf{X}}_{T-L:T+H}^{(ij)}, \ \ \tilde{\mathbf{S}}_{T-L:T+H}^{(ij)} \,\big),
\end{equation*}
where $\mathbf{\xi}_{T-L:T+H}^{(ij)} \in \mathbb{R}^{(L+H) \times (6d_{\mathrm{emb}}+5)}$ denotes the final input matrix.

\subsubsection{Feature Fusion}

Before feeding the input into the temporal modeling network, early feature fusion is performed to jointly encode different features including sailing durations, vessel counts in destination port, static features of vessels, segment features and time window identifiers within each time window. Early feature fusion is necessary for the extraction of cross features between sailing durations and other dynamic or static features, which can help explain variations in sailing durations.

Concretely, let $\mathbf{\xi}_{t}^{(ij)}$ obtained from Section~\ref{sec:input_representation} denote the input vector in time window $t$ for segment $\textcolor{black}{p^{(i)}} \rightarrow \textcolor{black}{p^{(j)}}$. The one-layer linear transformation is used to obtain the feature fusion representation:
\begin{equation*}
\mathbf{h}_{t}^{(ij)}=\mathbf{\xi}_{t}^{(ij)}\mathbf{W}_1\,+\mathbf{b}_1,
\end{equation*}
where $\mathbf{W}_1 \in \mathbb{R}^{(6d_{\mathrm{emb}}+5) \times d_{\mathrm{model}}}$ is the learnable transformation matrix, $b_1 \in \mathbb{R}^{d_{\mathrm{model}}}$ serves as the bias vector and $d_{\mathrm{model}}$ is the dimension for the temporal modeling network. Let ${\mathbf{H}}^{(ij)}$ represent $\big[{\mathbf{h}}_t^{(ij)}\big]_{t=T-L}^{T+(H-1)}$ and ${\mathbf{H}}^{(ij)}$ becomes the input to the downstream temporal modeling network.

\subsubsection{Temporal Modeling Network}
\label{sec:TMN}

To capture and infer dependencies across time windows, a transformer-based network is employed to model the temporal dynamics of the sailing durations on voyage segments. Absolute positional encoding ($\mathrm{PE}_t$) is added to the hidden vector $\mathbf{h}_t^{(ij)}$:
\begin{equation*}
\mathrm{PE}_t(2\textcolor{black}{r})=\sin\!\Big(t\,\omega_{\textcolor{black}{r}}\Big), 
\qquad
\mathrm{PE}_t(2\textcolor{black}{r}{+}1)=\cos\!\Big(t\,\omega_{\textcolor{black}{r}}\Big),
\qquad
\omega_{\textcolor{black}{r}} = 1000^{-\,\frac{2{\textcolor{black}{r}}}{d_{\mathrm{model}}}},
\end{equation*}
\begin{equation*}
\tilde{\mathbf{h}}_t^{(ij)} = \mathbf{h}_t^{(ij)} + \mathrm{PE}_t,
\end{equation*}
where hidden vectors in different time windows are assigned with a unique mark to indicate their positions. $\tilde{\mathbf{h}}_t^{(ij)}$ becomes the hidden vector with the positional mark.

\begin{figure}[h]
    \centering
    \includegraphics[width=0.95\textwidth]{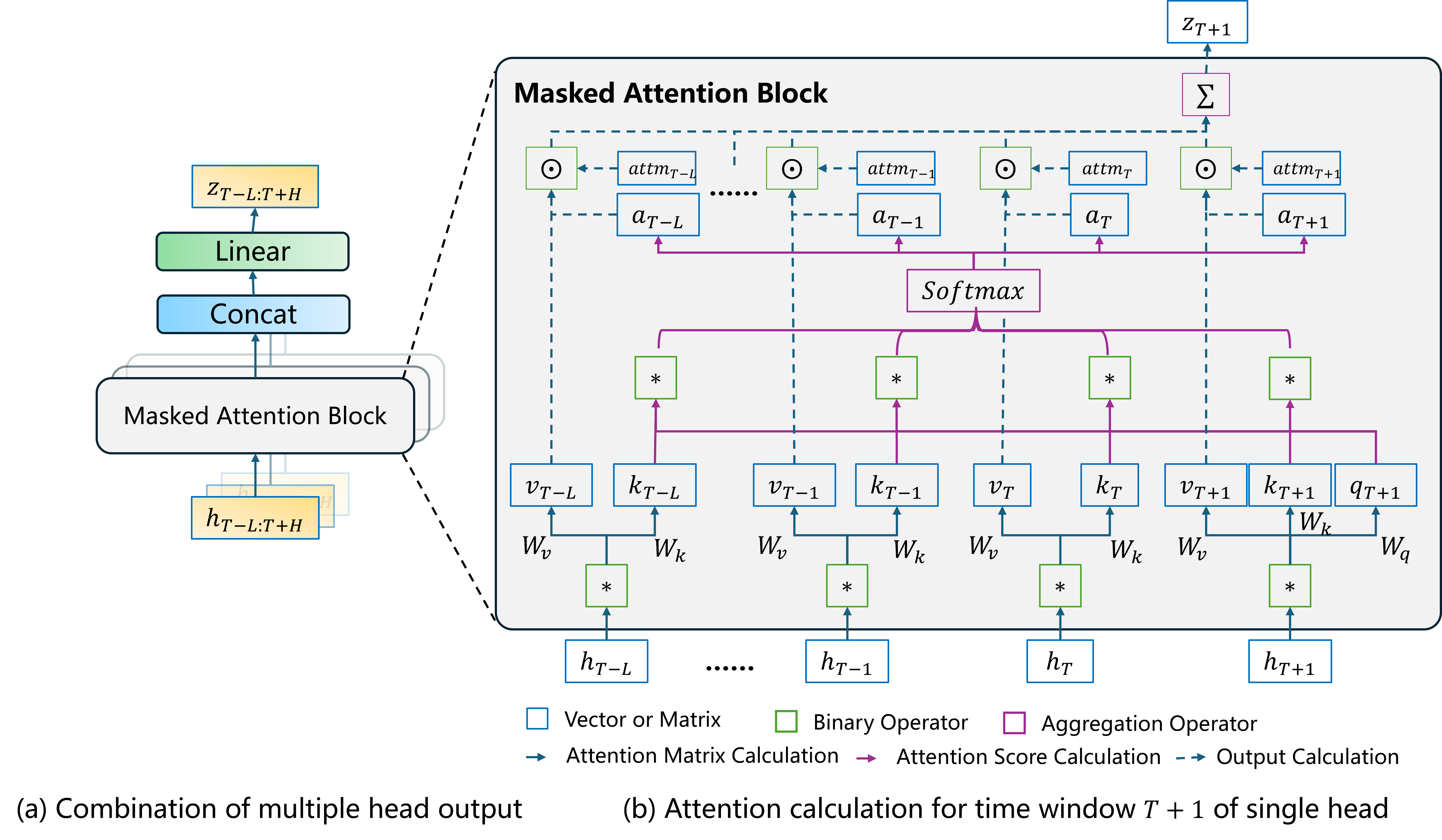}
    \caption{Sketch of masked multi-head attention mechanism.}
    \label{fig:mha}
\end{figure}

The attention mechanism forms the core computational unit of the transformer block (\cite{vaswani2023attentionneed}), enabling the model to dynamically weigh and aggregate relevant historical context when making predictions for any given time window. As illustrated in Figure~\ref{fig:mha}, for a target prediction step (e.g., time window \(T+1\)), each hidden vector \(\tilde{\mathbf{h}}_t^{(ij)}\) is linearly projected into three distinct vectors: a query \(\mathbf{q}_t^{(ij)}\), a key \(\mathbf{k}_t^{(ij)}\), and a value \(\mathbf{v}_t^{(ij)}\). The query for the current step is compared with all historical keys via scaled dot-product operations to compute a set of attention weights. These weights determine the relevance of each historical time step to the current prediction. The final contextual representation for the step is then obtained as a weighted sum of the corresponding value vectors. To capture diverse temporal dependencies, the multi-head attention mechanism is employed. Specifically, multiple sets of query, key, and value projections are learned in parallel, each operating in a reduced dimensional subspace. This allows the network to jointly attend to information from different representation subspaces. Formally, let $\tilde{\mathbf{H}}^{(ij)}$ represent $\big[\tilde{\mathbf{h}}_t^{(ij)}\big]_{t=T-L}^{T+(H-1)}$, the temporal modeling network follows the equations below:
\begin{equation*}
\mathbf{Q}_h^{(ij)} = \tilde{\mathbf{H}}^{(ij)}\mathbf{W}_h^{Q},\qquad
\mathbf{K}_h^{(ij)} = \tilde{\mathbf{H}}^{(ij)}\mathbf{W}_h^{K},\qquad
\mathbf{V}_h^{(ij)} = \tilde{\mathbf{H}}^{(ij)}\mathbf{W}_h^{V}, \qquad h=1,\dots,n_{\mathrm{head}}
\end{equation*}

\begin{equation*}
\mathbf{A}_h^{(ij)} \;=\; \frac{\mathbf{Q}_h^{(ij)} \mathbf{K}_h^{(ij)\top}}{\sqrt{d_{\mathrm{model}}\textcolor{black}{/n_{\mathrm{head}}}}} \; \odot\mathbf{AttMask} + ( \mathbf{1}-\mathbf{AttMask})\cdot(-\infty),\qquad
\boldsymbol{\alpha}_h^{(ij)} \;=\; \mathrm{Softmax}(\mathbf{A}_h^{(ij)}),
\end{equation*}

\begin{equation*}
\mathbf{Z}_h^{(ij)} \;=\; {\boldsymbol{\alpha}}_h^{(ij)} \mathbf{V}_h^{(ij)},\qquad
\mathbf{Z}^{(ij)} \;=\; \mathrm{Dropout}\!\left(\mathrm{Concat}\big(\mathbf{Z}_1^{(ij)},\cdots,\mathbf{Z}_{\textcolor{black}{n_{\mathrm{head}}}}^{(ij)}\,\big)\mathbf{W}^{O};\,p_{\mathrm{att}}\right).
\end{equation*}
$\mathbf{W}_h^{Q} \in \mathbb{R}^{d_{\mathrm{model}} \times (d_{\mathrm{model}}/n_{\mathrm{head}})}$, $\mathbf{W}_h^{K} \in \mathbb{R}^{d_{\mathrm{model}} \times (d_{\mathrm{model}}/n_{\mathrm{head}})}$, $\mathbf{W}_h^{V} \in \mathbb{R}^{d_{\mathrm{model}} \times (d_{\mathrm{model}}/n_{\mathrm{head}})}$, $\mathbf{W}^{O} \in \mathbb{R}^{d_{\mathrm{model}} \times d_{\mathrm{model}}}$ are learnable parameters. $\mathbf{Q}_h^{(ij)}$, $\mathbf{K}_h^{(ij)}$, $\mathbf{V}_h^{(ij)}$, $\boldsymbol{\alpha}_h^{(ij)}$ represent the query matrix, key matrix, value matrix and the attention matrix of $h$th head respectively. $\mathbf{AttMask}$ affecting on the attention matrix $\mathbf{A}_h$ denotes the causal mask to prevent information leakage from future time steps:

\begin{equation*}
\mathbf{AttMask}(t,s)=\begin{cases}
1,& s\le t,\\
0,& s>t,
\end{cases}
\qquad \mathbf{AttMask}\in\{0,1\}^{(L+H)\times (L+H)},
\end{equation*}
where $\mathbf{AttMask}$ is a lower triangular matrix. The $\mathrm{Softmax(\cdot)}$ operator ensures the summation of attention weights from the current step to historical steps is equal to 1. The $\mathrm{Dropout}(\cdot;p_{\mathrm{att}})$ operator with the dropout rate $p_{\mathrm{att}}$ is utilized to avoid overfitting. $\mathbf{Z}_h^{(ij)}$ and $\mathbf{Z}^{(ij)}$ are the output of the $h$th attention head and final output of masked multi-head attention mechanism respectively. To ensure training stability and effective gradient flow during backpropagation, residual connections and layer normalization are employed:

\begin{equation*}
\tilde{\mathbf{Z}}^{(ij)} = \mathrm{LayerNorm}(\mathbf{Z}^{(ij)} + \tilde{\mathbf{H}}^{(ij)}).
\end{equation*}
The $\mathrm{LayerNorm}(\cdot)$ operator normalizes the vector on each time step and $\tilde{\mathbf{Z}}^{(ij)}$ is the hidden matrix after normalization.

To further refine these representations and enhance the model's expressive power, a position-wise feed forward network (FFN) is applied to each temporal position independently. This sub-layer consists of two linear transformations with a ReLU activation in between, introducing a non-linear transformation that operates identically on each time step:

\begin{equation*}
\mathbf{O}^{(ij)} = \mathrm{Dropout}\big(\mathrm{ReLU}\!\left(\tilde{\mathbf{Z}}^{(ij)}\mathbf{W}_2 + \mathbf{b}_2\right);p_{\mathrm{ffn}}\big)\mathbf{W}_3 + \mathbf{b}_3,
\end{equation*}
\begin{equation*}
\tilde{\mathbf{O}}^{(ij)} = \mathrm{LayerNorm}(\mathbf{O}^{(ij)} + \tilde{\mathbf{Z}}^{(ij)}).
\end{equation*}
$\mathbf{W}_2 \in \mathbb{R}^{d_{\mathrm{model}} \times 4d_{\mathrm{model}}}$, $\mathbf{W}_3 \in \mathbb{R}^{4d_{\mathrm{model}} \times d_{\mathrm{model}}}$, $\mathbf{b}_2 \in \mathbb{R}^{4d_{\mathrm{model}}}$, $\mathbf{b}_3 \in \mathbb{R}^{d_{\mathrm{model}}}$ are learnable transformation matrices and biases respectively. Dropout mechanism with rate $p_{\mathrm{ffn}}$ is also employed to avoid overfitting. $\tilde{\mathbf{O}}^{(ij)}$ denotes the output of the FFN block. A residual connection followed by layer normalization is employed to stabilize and accelerate the training process.

The whole operation above in Section~\ref{sec:TMN} is defined as a Temporal Modeling Network Block (TMNBlock). The block can be built deeper to integrate hidden features:
\begin{equation*}
\tilde{\mathbf{O}}^{(ij)} = {\mathbf{TMNBlock}}^{n_{\mathrm{block}}}(\mathbf{H}^{(ij)}),
\end{equation*}
where $n_{\mathrm{block}}$ denotes the number of stacking TMNBlocks.

\subsubsection{Multi-Task Output Layer}

The output layer converts the hidden matrix $\tilde{\mathbf{O}}^{(ij)}$ into the two final output vectors: sailing durations and vessel counts in the destination port in future $H$ time steps. A two-layer MLP is used in this part:
\begin{equation*}
\big(\hat{\mathbf{Y}}_{T:T+H}^{(ij)}, \hat{\mathbf{X}}_{T:T+H}^{(ij)}\big) = [\mathrm{ReLU}\big({\tilde{\mathbf{O}}^{(ij)}}\mathbf{W}_4 + \mathbf{b}_4\big)\mathbf{W}_5+\mathbf{b}_5]_{T:T+H}.
\end{equation*}
$\mathbf{W}_4 \in \mathbb{R}^{d_{\mathrm{model}} \times d_{\mathrm{temp}}}$, $\mathbf{W}_5 \in \mathbb{R}^{d_{\mathrm{temp}} \times 2}$, $\mathbf{b}_4 \in \mathbb{R}^{d_{\mathrm{temp}}}$, $\mathbf{b}_5 \in \mathbb{R}^{2}$ are learnable transformation matrices and biases respectively and $d_{\mathrm{temp}}$ is the intermediate dimension. As for the final output of the model, only the latest $H$ steps of $\hat{\mathbf{Y}}^{(ij)}$ and $\hat{\mathbf{X}}^{(ij)}$ are selected to calculate for the loss on segment $\textcolor{black}{p^{(i)}} \rightarrow \textcolor{black}{p^{(j)}}$. Besides, the mask $\mathbf{M}_{T:T+H}^{(ij)}$ is constructed to avoid the loss calculation on missing values of sailing durations:
\begin{equation*}
\mathbf{M}_{t}^{(ij)}=\begin{cases}
1,& y_t^{(ij)}>0,\\
0,& \mathrm{otherwise},
\end{cases}
\end{equation*}
where $y_t^{(ij)}$ and $\mathbf{M}_{t}^{(ij)}$ denote the real sailing duration and the mask on segment $\textcolor{black}{p^{(i)}} \rightarrow \textcolor{black}{p^{(j)}}$ with departure timestamp in time window $t$ respectively.

The loss function for the main task of sailing time predictions integrates mean absolute error (MAE) and mean absolute percentage error (MAPE) because both absolute and relative deviations are important. A scalar $\beta\in[0,1]$ balances the two terms. Define the masked absolute error for segment $\textcolor{black}{p^{(i)}} \rightarrow \textcolor{black}{p^{(j)}}$ in time window $t$ as
\begin{equation*}
e_{t}^{(ij)} \;=\; \mathbf{M}_{t}^{(ij)}\,\big|\,y_{t}^{(ij)} - \hat{y}_{t}^{(ij)}\big|.
\end{equation*}
Given the time window $T$ to make predictions, the masked MAE and MAPE over the forecasting horizon $H$ on segment $\textcolor{black}{p^{(i)}} \rightarrow \textcolor{black}{p^{(j)}}$ are defined as follows:
\begin{equation*}
\mathrm{MAE}^{(ij)}_T \;=\; \frac{\sum_{t=T}^{T+(H-1)} e_{t}^{(ij)}}{H},
\end{equation*}
\begin{equation*}
\mathrm{MAPE}^{(ij)}_T \;=\; \frac{\sum^{T+(H-1)}_{t={T}} \mathbf{M}_{t}^{(ij)}\,\dfrac{\big|\,y_{t}^{(ij)} - \hat{y}_{t}^{(ij)}\big|}{y_{t}^{(ij)}}}{H}.
\end{equation*}
The loss for the main task is a combination of $\mathrm{MAE}^{(ij)}_T$ and $\mathrm{MAPE}^{(ij)}_T$:
\begin{equation*}
\mathcal{L}_{T,\mathrm{main}}^{(ij)} \;=\; \beta\,\mathrm{MAE}_T^{(ij)} \;+\; \big(1-\beta\big)\,\mathrm{MAPE}_T^{(ij)},
\end{equation*}
where $\mathcal{L}_{T,\mathrm{main}}^{(ij)}$ denotes the loss for the main task over the forecasting horizon $H$ on segment $\textcolor{black}{p^{(i)}} \rightarrow \textcolor{black}{p^{(j)}}$, given the time window $T$ to make predictions.

For the auxiliary task of the destination port vessel count prediction of each segment, the mean absolute error is adopted:
\begin{equation*}
\mathcal{L}_{T, \mathrm{auxiliary}}^{(ij)} \;=\; \frac{\sum^{T+(H-1)}_{t=T} \big|\,x_{t}^{(ij)} - \hat{x}_{t}^{(ij)}\big|}{H},
\end{equation*}
where $\mathcal{L}_{T,\mathrm{auxiliary}}^{(ij)}$ denotes the auxiliary loss over the forecasting horizon $H$ on segment $\textcolor{black}{p^{(i)}} \rightarrow \textcolor{black}{p^{(j)}}$, given the time window $T$ to make predictions.

The total loss is a combination of losses for main task and auxiliary task. To keep the model concentrating on the main task, the main loss is set with a higher weight:
\begin{equation*}
\mathcal{L}^{(ij)}_T \;=\; \eta \mathcal{L}_{T,\mathrm{main}}^{(ij)} + (1-\eta)\mathcal{L}_{T,\mathrm{auxiliary}}^{(ij)},
\end{equation*}
where $\mathcal{L}_{T}^{(ij)}$ denotes the total loss over the forecasting horizon $H$ on segment $\textcolor{black}{p^{(i)}} \rightarrow \textcolor{black}{p^{(j)}}$, given the time window $T$ to make predictions.

For a mini batch of $\mathcal{|B|}$ samples, the training objective is to minimize the average loss of the batch $\mathcal{B}$:
\begin{equation*}
\mathcal{L}_\mathcal{B} \;=\; \frac{1}{\mathcal{\mathcal{|B|}}}\sum_{(i,j,T) \in \mathcal{B}} \mathcal{L}_T^{(ij)},
\end{equation*}
where $\mathcal{L}_\mathcal{B}$ denotes the batch loss.

During training, back propagation mechanism is adopted to update all trainable parameters. Because the main and auxiliary tasks share the representation, their gradients jointly act on the shared features, thereby capturing collaborative signals, improving the joint modeling of future sailing durations and destination port vessel counts. Note that the mask $\mathbf{M}_{t}^{(ij)}$ is applied consistently in both the loss and its gradients, so that only observed sailing records contribute to learning, preventing noises from missing values.

\section{Experiment}
\label{sec:experiment}

\subsection{Dataset Description}
\label{sec:description}

The original datasets used in the experiment include AIS dataset, port geofence dataset and vessel information dataset. The AIS dataset covers the trajectories of vessels in 2021 worldwide with over 200 million data points. The port geofence dataset contains regions of thousands of ports worldwide. To make sure only container \textcolor{black}{vessels} are considered in the experiment, vessel information dataset is utilized to filter the valid IMOs of container \textcolor{black}{vessels} based on the type field. In total, 5008 container \textcolor{black}{vessels} are selected.

\begin{figure}[h]
    \centering
    \includegraphics[width=0.95\textwidth]{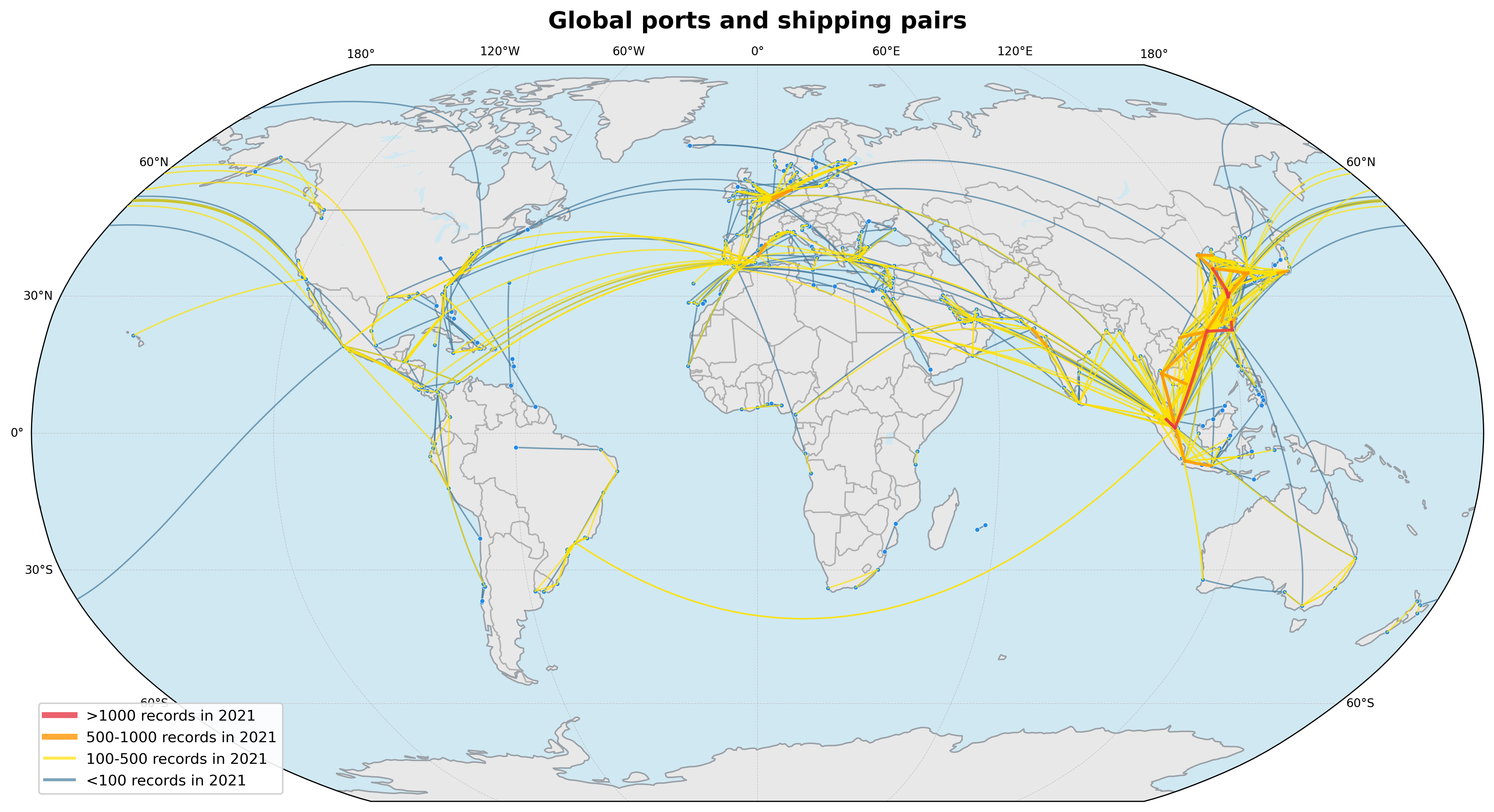}
    \caption{Distribution of global ports and shipping lines with different frequency.}
    \label{fig:segment}
\end{figure}

Through the data preprocessing pipeline described in Section~\ref{sec:process}, the vessel voyage record dataset and port vessel count dataset can be accessed for the experiment. The fields of vessel voyage record dataset are described in Table~\ref{tab:voyage_record}, which contain the sailing start time, the sailing vessel, the sailing voyage and the sailing duration. In the study, segments with more than 75 records in 2021 are considered because sparse data makes it difficult to capture the inherent patterns. Totally, 1120 segments and 380 ports are selected, which are shown in Figure~\ref{fig:segment}. Lines \textcolor{black}{represent} voyage records \textcolor{black}{occurred} between the two ports in 2021. It can be seen that most segments gather in the west coast of the Pacific and Europe with short to medium sailing distance. The most frequent segment is from Waigaoqiao port to Ningbo port, with 2310 observations and 5.37 hours as average sailing durations. The selected segments in the experiment cover 10.4\% of segments all over the world, including 54.5\% sailing records in 2021 with 201510 observations. It can be observed that medium-to-low-frequency segments and medium-to-short distance segments account for the majority. The frequency of segments usually becomes low \textcolor{black}{as segment distance increases}. As for the port vessel count dataset, it includes 380 ports with relative vessel counts on each time window $t$.

\begin{table}[h]
\centering
\caption{Fields and descriptions of regularized vessel voyage record dataset.}
\label{tab:voyage_record}
\renewcommand{\arraystretch}{1.15}
\begin{tabularx}{\textwidth}{lX}
\toprule
\textbf{Column name} & \textbf{Description} \\
\midrule
time window & Vessel departure time window \\ 
IMO & International Maritime Organization number (unique vessel identifier) \\ 
width & Vessel width (beam) \\ 
length & Vessel length \\ 
TEU & Maximum container capacity (Twenty-foot Equivalent Units) \\ 
crrName & Carrier company name \\ 
startPortName & Departure port name \\ 
endPortName & Destination port name \\ 
tmnName & Terminal name \\ 
ATA & Actual time of arrival \\ 
duration & Actual sailing duration \\
\bottomrule
\end{tabularx}
\end{table}

\subsection{Experiment Settings}

A comprehensive evaluation of the proposed model is conducted. The dataset is divided into a training set from January 1st, 2021 to October 31st, 2021 for model training and a test set from November 1st, 2021 to December 31st, 2021 \textcolor{black}{in chronological order} for model evaluation. During training, data from September 1st, 2021 to October 31st, 2021 is reserved as the validation set to choose the best model. Considering that it is inconvenient for management and updating to train a model for each segment, one model is trained to give predictions for all segments.

\begin{figure}[h]
    \centering
    \includegraphics[width=0.95\textwidth]{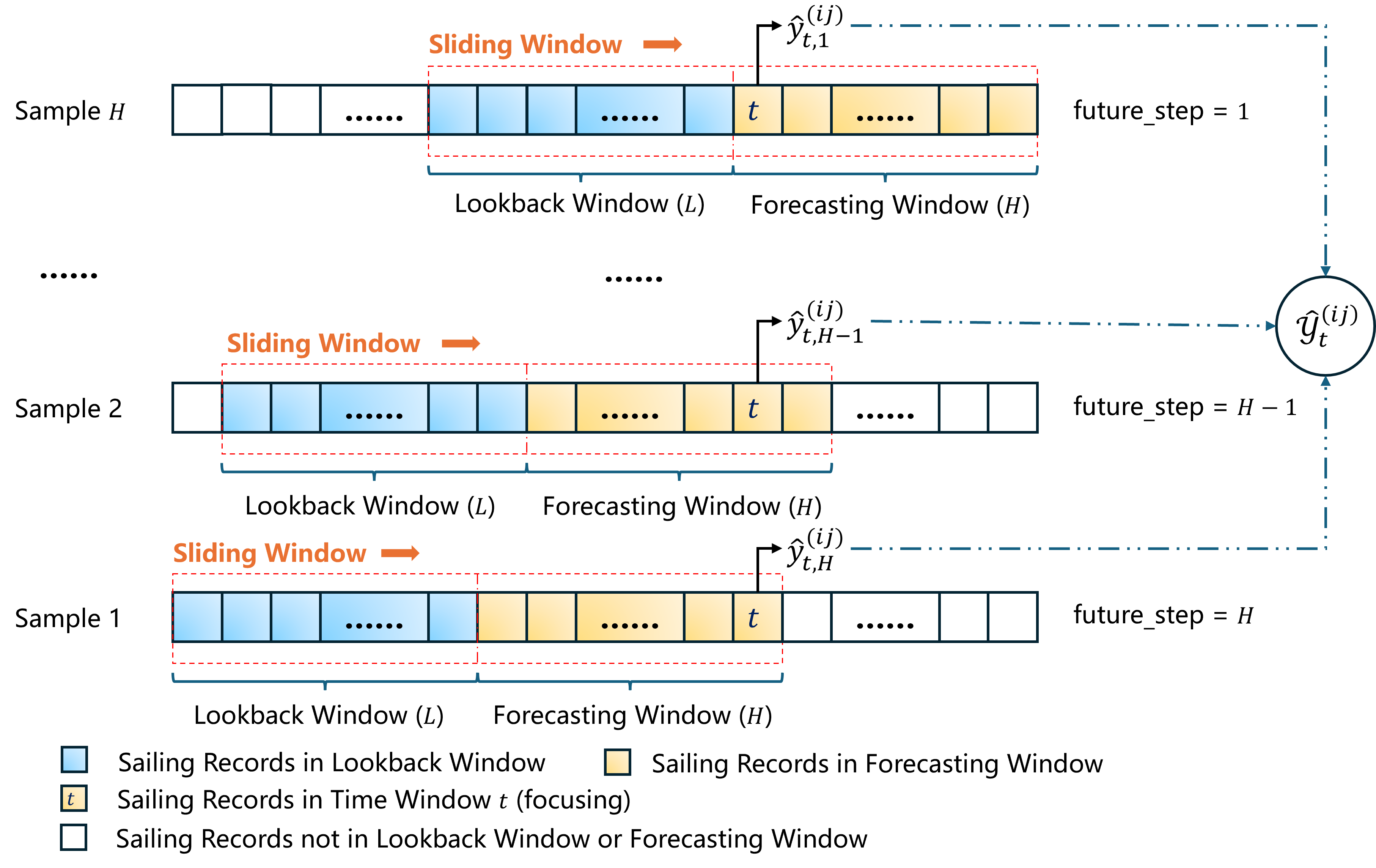}
    \caption{Sliding window mechanism for sample generation.}
    \label{fig:Sliding_window}
\end{figure}

The training, validation and test samples are generated from the vessel voyage record dataset and port vessel count dataset with a sliding window mechanism, as shown in Figure~\ref{fig:Sliding_window}. The voyage record dataset is ordered by time window for each segment respectively. The window with width $L+H$ slides along the time axis with the step one. The \textcolor{black}{first} $L$ observations serve as the lookback window while the \textcolor{black}{following} $H$ observations serve as the forecasting target. A sample is generated with the window sliding one step.

To better evaluate the performance of the model, several performance metrics are selected. Owing to the sliding window sample generation, for a given sailing record with duration $y_t^{(ij)}$, there can be up to $H$ distinct prediction results issued at different future steps for time-series models, which is shown in Figure~\ref{fig:Sliding_window}. Let $\hat{y}_{t, k}^{(ij)}$ denote the prediction result for $y_t^{(ij)}$ at future step $k$ and $\hat{\mathcal{Y}}_{t}^{(ij)} = \{\hat{y}_{t, 1}^{(ij)}, \hat{y}_{t, 2}^{(ij)}, \dots, \hat{y}_{t, \mathcal{K}}^{(ij)}\}$ denote the set of all prediction results for $y_t^{(ij)}$ at different future steps where $\mathcal{K} \leq H$. For non-time-series models, the prediction result $\hat{y}_{t, k}^{(ij)}$ at different future steps $k$ keeps the same. Therefore, the prediction error for the sailing record in test dataset is defined as the average over all its available predictions:
\begin{equation*}
\mathrm{MAE}_{t}^{(ij)} = \frac{1}{\mathcal{K}}\sum_{k=1}^{\mathcal{K}}\bigl|y_{t}^{(ij)} - \hat{y}_{t,k}^{(ij)}\bigr|,
\end{equation*}
\begin{equation*}
\mathrm{MAPE}_{t}^{(ij)} = \frac{1}{\mathcal{K}}\sum_{k=1}^{\mathcal{K}}\frac{\bigl|y_{t}^{(ij)} - \hat{y}_{t,k}^{(ij)}\bigr|}{y_{t}^{(ij)}},
\end{equation*}
\begin{equation*}
\textcolor{black}{\mathrm{RMSE}_{t}^{(ij)} = \sqrt{\frac{1}{\mathcal{K}}\sum_{k=1}^{\mathcal{K}}\bigl(y_{t}^{(ij)} - \hat{y}_{t,k}^{(ij)}\bigr)^2},}
\end{equation*}
where $\mathrm{MAE}_{t}^{(ij)}$, $\mathrm{MAPE}_{t}^{(ij)}$ \textcolor{black}{and $\mathrm{RMSE}_{t}^{(ij)}$} denote the predicted MAE, MAPE \textcolor{black}{and RMSE} of the sailing record departing in time window $t$ on segment $\textcolor{black}{p^{(i)}} \rightarrow \textcolor{black}{p^{(j)}}$.

Let $v^{(ij)}$ denote the number of sailing records in the test set on segment $\textcolor{black}{p^{(i)}} \rightarrow \textcolor{black}{p^{(j)}}$, the segment‑level error is then obtained by averaging errors of all test records belonging to that segment:
\begin{equation*}
\mathrm{MAE}^{(ij)} = \frac{1}{v^{(ij)}}\sum_{y^{(ij)}_t \in\ \mathrm{TestSet}}\mathrm{MAE}_{t}^{(ij)},
\end{equation*}
\begin{equation*}
\mathrm{MAPE}^{(ij)} = \frac{1}{v^{(ij)}}\sum_{y^{(ij)}_t \in\ \mathrm{TestSet}}\mathrm{MAPE}_{t}^{(ij)},
\end{equation*}
\begin{equation*}
\textcolor{black}{\mathrm{RMSE}^{(ij)} = \sqrt{\frac{1}{v^{(ij)}}\sum_{y^{(ij)}_t \in\ \mathrm{TestSet}}{\mathrm{RMSE}_{t}^{(ij)}}^2}},
\end{equation*}
where $\mathrm{MAE}^{(ij)}$, $\mathrm{MAPE}^{(ij)}$ \textcolor{black}{and $\mathrm{RMSE}^{(ij)}$} denote the predicted MAE, MAPE \textcolor{black}{and RMSE} on segment $\textcolor{black}{p^{(i)}} \rightarrow \textcolor{black}{p^{(j)}}$.

Given a collection of segments $\mathcal{S}$, the overall unweighted error is the arithmetic mean of the segment‑level errors; the record‑weighted error accounts for the number of test samples in each segment:
\begin{equation*}
\mathrm{MAE}^{\mathcal{S}}_{\mathrm{unweighted}} = \frac{1}{|\mathcal{S}|}\sum_{e^{(ij)} \in\ \mathcal{S}}\mathrm{MAE}^{(ij)},\ 
\mathrm{MAE}^{\mathcal{S}}_{\mathrm{weighted}} = \frac{1}{\displaystyle\sum_{e^{(ij)} \in \mathcal{S}} v^{(ij)}}\sum_{e^{(ij)} \in\ \mathcal{S}}v^{(ij)}\times \mathrm{MAE}^{(ij)},
\end{equation*}
\begin{equation*}
\mathrm{MAPE}^{\mathcal{S}}_{\mathrm{unweighted}} = \frac{1}{|\mathcal{S}|}\sum_{e^{(ij)} \in\ \mathcal{S}}\mathrm{MAPE}^{(ij)},\ 
\mathrm{MAPE}^{\mathcal{S}}_{\mathrm{weighted}} = \frac{1}{\displaystyle\sum_{e^{(ij)} \in \mathcal{S}} v^{(ij)}}\sum_{e^{(ij)} \in\ \mathcal{S}}v^{(ij)}\times \mathrm{MAPE}^{(ij)},
\end{equation*}
\begin{equation*}
\textcolor{black}{\mathrm{RMSE}^{\mathcal{S}}_{\mathrm{unweighted}} = \sqrt{\frac{1}{|\mathcal{S}|}\sum_{e^{(ij)} \in\ \mathcal{S}}{\mathrm{RMSE}^{(ij)}}^2},\ 
\mathrm{RMSE}^{\mathcal{S}}_{\mathrm{weighted}} = \sqrt{\frac{1}{\displaystyle\sum_{e^{(ij)} \in \mathcal{S}} v^{(ij)}}\sum_{e^{(ij)} \in\ \mathcal{S}}v^{(ij)}\times {\mathrm{RMSE}^{(ij)}}^2},}
\end{equation*}
where $\mathrm{MAE}^{\mathcal{S}}_{\mathrm{unweighted}}$, $\mathrm{MAPE}^{\mathcal{S}}_{\mathrm{unweighted}}$ and \textcolor{black}{$\mathrm{RMSE}^{\mathcal{S}}_{\mathrm{unweighted}}$} denote the predicted MAE, MAPE \textcolor{black}{and RMSE} on segment set $\mathcal{S}$ under the unweighted setting and $\mathrm{MAE}^{\mathcal{S}}_{\mathrm{weighted}}$, $\mathrm{MAPE}^{\mathcal{S}}_{\mathrm{weighted}}$ \textcolor{black}{and $\mathrm{RMSE}^{\mathcal{S}}_{\mathrm{weighted}}$} denote the predicted MAE, MAPE \textcolor{black}{and RMSE} on segment set $\mathcal{S}$ under the weighted setting. Unless otherwise stated, the predicted MAE, MAPE \textcolor{black}{and RMSE} for the set of voyage segment set $\mathcal{S}$ are calculated under the weighted version.

Besides, to analyze the error at different future steps over the forecast horizon, only those records with $H$ available prediction results are considered ($|\hat{\mathcal{Y}}_{t}^{(ij)}| = H$). Let $\mathcal{T}_{\mathrm{test}}^{(ij)}$ denote the set of time window indices of those considered sailing records for segment $e^{(ij)}$:
\[
\mathcal{T}_{\mathrm{test}}^{(ij)}
= \left\{\, t \;\middle|\; y_t^{(ij)} \in \mathrm{TestSet},\ \left|\hat{\mathcal{Y}}_{t}^{(ij)}\right| = H \,\right\}.
\]
The average error at the $k$th future step ($k = 1,\dots,H$) in the forecasting horizon is computed as:
\begin{equation*}
\mathrm{MAE}(k) = \frac{1}{\sum_{e^{(ij)} \in \mathcal{E}}\big|\mathcal{T}_{\mathrm{test}}^{(ij)}\big|}\sum_{e^{(ij)} \in \mathcal{E}}\sum_{t \in \mathcal{T}_{\mathrm{test}}^{(ij)}} \bigl|y_{t}^{(ij)} - \hat{y}_{t,k}^{(ij)}\bigr|,
\end{equation*}
\begin{equation*}
\mathrm{MAPE}(k) = \frac{1}{\sum_{e^{(ij)} \in \mathcal{E}}\big|\mathcal{T}_{\mathrm{test}}^{(ij)}\big|}\sum_{e^{(ij)} \in \mathcal{E}}\sum_{t \in \mathcal{T}_{\mathrm{test}}^{(ij)}} \frac{\bigl|y_{t}^{(ij)} - \hat{y}_{t,k}^{(ij)}\bigr|}{y_{t}^{(ij)}},
\end{equation*}
\begin{equation*}
\textcolor{black}{\mathrm{RMSE}(k) = \sqrt{\frac{1}{\sum_{e^{(ij)} \in \mathcal{E}}\big|\mathcal{T}_{\mathrm{test}}^{(ij)}\big|}\sum_{e^{(ij)} \in \mathcal{E}}\sum_{t \in \mathcal{T}_{\mathrm{test}}^{(ij)}} \bigl(y_{t}^{(ij)} - \hat{y}_{t,k}^{(ij)}\bigr)^2},}
\end{equation*}
where $\mathrm{MAE}(k)$, $\mathrm{MAPE}(k)$ \textcolor{black}{and $\mathrm{RMSE}(k)$} denote the predicted MAE, MAPE \textcolor{black}{and RMSE} on future step $k$. This step‑wise error profile reveals how predictive accuracy evolves with the future time step increasing.

\textcolor{black}{The experiments are conducted on a Linux server (Ubuntu 22.04) equipped with two Intel Xeon Platinum 8358 CPUs (2.6 GHz, 32 cores per CPU), 2 TB DDR4 memory, and NVIDIA Tesla A100 GPUs with 80 GB memory. The software environment includes Python 3.10 with PyTorch, under CUDA 12.4 and NVIDIA driver version 550.54.15.}

\textcolor{black}{Hyperparameter settings of the experiment and the proposed model are decided as follows. For regularization, dropout rates of the model $p_{\mathrm{ffn}}$, $p_{\mathrm{att}}$ are set to 0.1. For optimization and training settings, Adam optimizer is utilized with default parameters; learning rate is set to $3e^{-3}$ initially and halved every 10 epochs; batch size is $1024$ and at most 30 epochs are used for training. For the task settings, the time resolution $\Delta t$ is set to 6 hours with lookback window length $L$ of 168 time windows (42 days) and forecasting horizon $H$ of 84 time windows (21 days). Loss weight of MAE $\beta$ is set to 0.8, confirming that MAE loss and MAPE loss are comparable. Loss weight for main task $\eta$ is set to 0.9. For the model architecture, a grid-search experiment is conducted to find the best hyperparameter combination. Concretely, the number of stacking TMNBlocks $n_{\mathrm{block}}$, the embedding dimension $d_{\mathrm{emb}}$ and the model dimension $d_{\mathrm{model}}$ are changed separately. The number of attention heads $n_{\mathrm{head}}$ is set to $d_{\mathrm{model}} / 4$ and the intermediate dimension of multi-task output layer $d_{\mathrm{temp}}$ is set to $d_{\mathrm{model}} / 2$. Totally 8 configurations are considered, covering different combinations of network depth and representation size. As reported in Table~\ref{tab:capacity_sensitivity}, taking both MAE, MAPE and RMSE into consideration, the model with 2 stacking TMNBlocks, embedding dimension of 32 and model dimension of 32 achieves the best comprehensive performance across the three evaluation metrics. Finally, the total hyperparameter settings  of the experiment are shown in Table~\ref{tab:hyperparams}. Furthermore, to examine the stability of the proposed framework, we conduct repeated experiments with different random seeds. As shown in Appendix~\ref{sec:repeat_experiments}, the proposed model achieves highly consistent results, with standard deviations of only 0.021 h, 0.006\%, and 0.073 h for MAE, MAPE, and RMSE, respectively. This confirms the repeatability and robustness of the proposed framework under stochastic ship navigation data.}

\begin{table}[htbp]
\centering
\caption{\textcolor{black}{Sensitivity analysis with respect to model capacity.}}
\label{tab:capacity_sensitivity}
\begin{tabularx}{\textwidth}{XXl|lll}
\toprule
\multicolumn{3}{c}{\textbf{Model Configuration}} & \multicolumn{3}{c}{\textbf{Performance Metrics}} \\
\cmidrule(r){1-3} \cmidrule(l){4-6}
\textbf{\#TMNBlocks} & \textbf{Embedding Dimension} & \textbf{Model Dimension} & \textbf{MAE (h)} & \textbf{MAPE} & \textcolor{black}{\textbf{RMSE (h)}} \\
\midrule
1 & 16 & 32 & 6.15 & 18.23\% & \textcolor{black}{12.06} \\
1 & 16 & 64 & 6.14 & 19.15\% & \textcolor{black}{12.14} \\
1 & 32 & 32 & 6.15 & 18.21\% & \textcolor{black}{12.09} \\
1 & 32 & 64 & 6.15 & 18.73\% & \textcolor{black}{12.10} \\
2 & 16 & 32 & 6.17 & \underline{18.10\%} & \textcolor{black}{12.32} \\
2 & 16 & 64 & \textbf{6.07} & 18.91\% & \textcolor{black}{\textbf{12.03}} \\
2 & 32 & 32 & \underline{6.08} & \textbf{18.05\%} & \textcolor{black}{\underline{12.04}} \\
2 & 32 & 64 & 6.17 & 18.37\% & \textcolor{black}{12.11} \\
\bottomrule
\end{tabularx}
\end{table}

\begin{table}[h]
\centering
\caption{Hyperparameter settings of the experiment and the proposed model.}
\label{tab:hyperparams}
\renewcommand{\arraystretch}{1.15}
\begin{tabularx}{\textwidth}{llXX}
\toprule
\textbf{Category} & \textbf{Hyperparameter} & \textbf{Symbol} & \textbf{Value} \\
\midrule
\multirow{5}{*}{Architecture} 
& Embedding dimension        & $d_{\mathrm{emb}}$                 & 32 \\
& Model dimension        & $d_{\mathrm{model}}$                 & 32 \\
& Number of stacking TMNBlocks                 & $n_{\mathrm{block}}$                 & 2 \\
& Number of attention heads                & $n_{\mathrm{head}}$                 & 8 \\
& Intermediate dimension of output layer                & $d_{\mathrm{temp}}$                 & 16 \\
\midrule
\multirow{2}{*}{Regularization} 
& Attention dropout rate                   & $p_{\mathrm{att}}$ & 0.1 \\
& FFN dropout rate                         & $p_{\mathrm{ffn}}$  & 0.1 \\
\midrule
\multirow{5}{*}{Optimization \& Training} 
& Optimizer                                & --                & Adam \\
& Learning rate                            & --              & $3e^{-3}$ \\
& LR schedule                              & --                  & step \\
& Batch size                               & $\mathcal{|B|}$                 & 1024 \\
& Max epochs                               & --                & 30 \\
\midrule
\multirow{4}{*}{Task Setup} 
& Lookback window length                  & $L$                 & 168 \\
& Forecast horizon length                         & $H$                 & 84 \\
& Time resolution                          & $\Delta t$          & 6 h \\
& Loss weight for MAE                      & $\beta$             & 0.8 \\
& Loss weight for main task                      & $\eta$             & 0.9 \\
\bottomrule
\end{tabularx}
\end{table}

\subsection{Result Analysis}

To better describe the prediction performance of the proposed model, some baseline models are selected to compare, which have been demonstrated to show strong performance from the previous study. Due to baseline models containing time series models and non-time-series models, historical time series are removed from the input and the input data is reconstructed as tabular data for non-time-series models. The baseline models are shown as follows:
\begin{itemize}
  \item \textbf{Light gradient boosting machine (LightGBM)} \citep{LEI2024116838}. A gradient boosting decision tree framework with histogram-based splitting and leaf-wise growth, offering strong accuracy and efficiency on large-scale tabular data.
  \item \textbf{Categorical boosting (CatBoost)}. Gradient boosting that natively handles categorical variables via ordered target statistics and permutation-driven training.
  \item \textbf{Extreme gradient boosting (XGBoost)} \citep{LEI2024116838}. Regularized gradient boosted trees with shrinkage and column subsampling; a strong tabular model known for robust performance.
  \item \textbf{Stacking model (LightGBM + CatBoost + XGBoost + Linear regression)} \citep{CHU2025105128}. An ensemble that blends the predictions of three boosted-tree learners using a linear meta-learner.
  \item \textbf{Long short-term memory (LSTM)} \citep{bulk}. A recurrent neural network with gating mechanisms to capture temporal dependencies and long-range context in sequence data.
  \item \textbf{iTransformer} \citep{liu2024itransformerinvertedtransformerseffective}. A transformer-based time-series model that treats variables as tokens and time as channels, enabling global cross-variable attention.
  \textcolor{black}{\item \textbf{Autoformer} \citep{wu2022autoformerdecompositiontransformersautocorrelation}. A Transformer-based model for long-term time series forecasting that employs decomposition architecture and autocorrelation-based attention to capture periodic patterns efficiently.}
  \textcolor{black}{\item \textbf{FEDformer} \citep{zhou2022fedformerfrequencyenhanceddecomposed}. A frequency-enhanced Transformer that performs forecasting in the frequency domain, combining Fourier or wavelet transforms with attention to improve long-horizon prediction accuracy.}
\end{itemize}
\textcolor{black}{For the gradient boosting baselines, including LightGBM, XGBoost, and CatBoost, hyperparameters were tuned using Optuna. Specifically, for each model, 20 optimization trials were conducted under the same training-validation split described above. In each trial, a candidate hyperparameter configuration was evaluated. The configuration achieving the best comprehensive performance across the three evaluation metrics was selected as the final setting for that model. For the stacking baseline, no separate hyperparameter tuning was performed, since the hyperparameter settings of its base learners were directly inherited from those of the corresponding individually trained base models. The meta-learner was set as linear regression. For the sequential deep learning models, the grid search method is utilized to find the optimal hyperparameter combinations. The hyperparameter combination with the best comprehensive performance in the validation set was selected. The key tuned hyperparameters and the final selected configurations are reported in Appendix~\ref{sec:hp}.}

Table~\ref{tab:results} shows that the proposed model delivers the best performance under both record-weighted and unweighted evaluations. In the weighted setting, it attains \textbf{6.08} h MAE and \textbf{18.05\%} MAPE, \textcolor{black}{and \textbf{12.04} h RMSE}. \textcolor{black}{Compared with the best baseline MAE achieved by Autoformer (6.38 h), the proposed model yields a relative reduction of \textbf{4.70\%} in MAE.} Compared with the best baseline MAPE \textcolor{black}{and RMSE} achieved by LSTM (18.99\%, \textcolor{black}{12.36 h}), it achieves relative reductions of \textbf{4.95\%} in MAPE \textcolor{black}{and \textbf{2.59\%} in RMSE}. Under the unweighted setting, the proposed model achieves \textbf{7.23} h MAE, \textbf{17.21\%} MAPE, \textcolor{black}{and \textbf{15.60} h RMSE, surpassing FEDformer (\textbf{7.56} h, \textbf{24.64\%}, \textbf{15.70} h) with a relative reduction of \textbf{4.37\%} in MAE and \textbf{0.64\%} in RMSE,} and outperforming LSTM (\textbf{7.60} h, \textbf{18.24\%}, \textcolor{black}{\textbf{15.87} h}) with a relative reduction of \textbf{5.65\%} in MAPE. Among the tree-based baselines, Stacking achieves the lowest MAE and RMSE, while LightGBM achieves the lowest MAPE. \textcolor{black}{Compared with these strongest tree-based results, the proposed model reduces MAE by \textbf{7.03\%} and \textbf{5.86\%}, and reduces MAPE by \textbf{39.49\%} and \textbf{38.58\%} in the weighted and unweighted settings, respectively.} These consistent margins indicate that the proposed architecture of early feature fusion and attention across time windows with auxiliary task learning captures long-horizon temporal patterns more effectively than both tree-based methods and representative deep-learning baselines. \textcolor{black}{The relatively weaker performance of iTransformer may be attributed to the fact that its vanilla architecture is primarily designed for multivariate time-series forecasting and does not explicitly incorporate a dedicated mechanism for fusing continuous temporal features with heterogeneous auxiliary attributes, such as static vessel features \citep{wang2024timexerempoweringtransformerstime}. Similarly, the Autoformer and FEDformer models, which have been proposed for long-horizon time series forecasting, achieved MAE and RMSE values close to LSTM but relatively weaker performance on MAPE. Autoformer leverages a decomposition architecture and autocorrelation-based attention to capture periodic patterns efficiently, while FEDformer operates in the frequency domain to enhance long-term prediction accuracy. However, both models are primarily designed for standard time series without extensive missing values or heterogeneous auxiliary features. In the ETA prediction task, missing values of sailing durations and exogenous variables such as vessel attributes pose a significant challenge for these architectures, as they lack dedicated mechanisms to integrate such heterogeneous inputs effectively.} The large MAPE gaps of the proposed model versus tree methods further suggest improved robustness to scale heterogeneity across segments, and the alignment of gains in weighted and unweighted views implies the benefits are not confined to high-frequency segments but generalize across the segment distribution.

\begin{table}[h]
\centering
\caption{\textcolor{black}{Performance comparison of models on weighted and unweighted settings.}}
\label{tab:results}
\renewcommand{\arraystretch}{1.15}
\begin{tabularx}{\textwidth}{lXXXXXX}
\toprule
\multirow{2}{*}{\textbf{Model}} & \multicolumn{3}{c}{\textbf{Weighted setting}} & \multicolumn{3}{c}{\textbf{Unweighted setting}} \\
\cmidrule(lr){2-4}\cmidrule(lr){5-7}
 & \textbf{MAE (h)} & \textbf{MAPE} & \textcolor{black}{\textbf{RMSE (h)}} & \textbf{MAE (h)} & \textbf{MAPE} & \textcolor{black}{\textbf{RMSE (h)}}\\
\midrule
\textbf{Our Model} & \textbf{6.08} & \textbf{18.05\%} & \textcolor{black}{\textbf{12.04}} & \textbf{7.23} & \textbf{17.21\%} & \textcolor{black}{\textbf{15.60}}\\
LightGBM     & \textcolor{black}{6.62} & \textcolor{black}{29.83\%} & \textcolor{black}{13.95} & \textcolor{black}{7.90} & \textcolor{black}{28.02\%} & \textcolor{black}{17.99}\\
CatBoost     & \textcolor{black}{7.69} & \textcolor{black}{42.22\%} & \textcolor{black}{15.57} & \textcolor{black}{9.30} & \textcolor{black}{39.45\%} & \textcolor{black}{19.81}\\
XGBoost      & \textcolor{black}{7.40} & \textcolor{black}{54.40\%} & \textcolor{black}{14.01} & \textcolor{black}{8.64} & \textcolor{black}{45.73\%} & \textcolor{black}{17.58}\\
Stacking     & \textcolor{black}{6.54} & \textcolor{black}{37.19\%} & \textcolor{black}{12.59} & \textcolor{black}{7.68} & \textcolor{black}{33.53\%} & \textcolor{black}{15.82}\\
LSTM         & 6.39 & 18.99\% & \textcolor{black}{12.36} & 7.60 & 18.24\% & \textcolor{black}{15.87}\\
iTransformer & 6.52 & 20.70\% & \textcolor{black}{12.45} & 7.77 & 19.64\% & \textcolor{black}{15.90}\\
\textcolor{black}{Autoformer} & \textcolor{black}{6.38} & \textcolor{black}{23.81\%} & \textcolor{black}{12.42} & \textcolor{black}{7.58} & \textcolor{black}{24.71\%} & \textcolor{black}{15.74}\\
\textcolor{black}{FEDformer} & \textcolor{black}{6.39} & \textcolor{black}{23.78\%} & \textcolor{black}{12.40} & \textcolor{black}{7.56} & \textcolor{black}{24.64\%} & \textcolor{black}{15.70}\\
\bottomrule
\end{tabularx}
\end{table}

Table~\ref{tab:all_bins} reports the forecasting results after grouping segments by their \emph{average sailing durations}. The proposed model consistently achieves the best MAE and MAPE across all four duration bins \textcolor{black}{while keeping competitive performance on RMSE}. In short segments ($<24$ h), it reduces MAE from \textbf{2.24} h to \textbf{2.21} h (\textbf{1.34\%} relatively) and MAPE from \textbf{23.45\%} to \textbf{22.51\%} (\textbf{4.01\%} relatively), \textcolor{black}{with a little margin compared with the best baseline model on RMSE}. For segments with average sailing durations 24--72 h, MAE drops from \textcolor{black}{\textbf{7.16} h} to \textbf{6.89} h (\textcolor{black}{\textbf{3.77\%}} relatively), MAPE from \textbf{17.08\%} to \textbf{16.18\%} (\textbf{5.27\%} relatively), \textcolor{black}{and RMSE from \textbf{11.31} h to \textbf{11.10} h (\textbf{1.86\%} relatively)}. For segments with average sailing durations 72--168 h, the proposed model again ranks first on all three metrics, with MAE reduced from \textcolor{black}{\textbf{12.42} h} to \textbf{11.97} h (\textcolor{black}{\textbf{3.62\%}} relatively), MAPE reduced from \textbf{12.35\%} to \textbf{11.50\%} (\textbf{6.88\%} relatively), \textcolor{black}{and similar RMSE performances with the best baseline}. In long segments (168--504 h), the proposed model achieves a reduction of MAE from \textcolor{black}{\textbf{27.05} h} to \textbf{26.92} h (\textcolor{black}{\textbf{0.48\%}} relatively) and MAPE from \textbf{11.55\%} to \textbf{10.94\%} (\textbf{5.28\%} relatively), \textcolor{black}{with a relatively higher RMSE}. These gains are most pronounced on segments with medium sailing durations because the record data is dense and regular, allowing the attention layers to capture the changing patterns of time series. The horizon from the predicting time to vessel arrival time at destination port spans from days to several weeks, allowing the auxiliary task to reduce the uncertainty of port congestion levels in the future, providing a side signal that sharpens sailing-duration estimations. Tree-based learners remain substantially worse in MAPE across all bins, reinforcing that explicit temporal modeling is critical. Practically, the result suggests prioritizing the proposed model for medium length segments to improve berth planning and rotations.

\begin{table}[h]
\centering
\caption{\textcolor{black}{Performance comparison of models for different segment groups binned by average sailing duration.}}
\label{tab:all_bins}
\renewcommand{\arraystretch}{1.12}
\begin{threeparttable}
\begin{tabularx}{\textwidth}{lllXlll}
\toprule
\textbf{Avg Dur}\tnote{a} & \textbf{\#Segments}\tnote{b} & \textbf{Sum}\tnote{c} & \textbf{Model} & \textbf{MAE (h)} & \textbf{MAPE} & \textcolor{black}{\textbf{RMSE (h)}}\\
\midrule
\multirow{9}{*}{\,$<$ 24h} & \multirow{9}{*}{406} & \multirow{9}{*}{86{,}455}
& \textbf{Our Model} & \textbf{2.21} & \textbf{22.51\%} & \textcolor{black}{4.87} \\
&&& LightGBM & \textcolor{black}{2.60} & \textcolor{black}{48.75\%} & \textcolor{black}{5.53}\\
&&& CatBoost & \textcolor{black}{3.22} & \textcolor{black}{76.03\%} & \textcolor{black}{6.81}\\
&&& XGBoost & \textcolor{black}{3.64} & \textcolor{black}{104.35\%} & \textcolor{black}{7.25}\\
&&& Stacking & \textcolor{black}{2.73} & \textcolor{black}{65.60\%} & \textcolor{black}{5.44}\\
&&& LSTM & 2.24 & 23.45\% & \textcolor{black}{4.84}\\
&&& iTransformer & 2.32 & 26.61\% & \textcolor{black}{\textbf{4.72}}\\
&&& \textcolor{black}{Autoformer} & \textcolor{black}{2.28} & \textcolor{black}{31.73\%} & \textcolor{black}{4.80}\\
&&& \textcolor{black}{FEDformer} & \textcolor{black}{2.28} & \textcolor{black}{31.57\%} & \textcolor{black}{4.82}\\
\midrule
\multirow{9}{*}{24h--72h} & \multirow{9}{*}{484} & \multirow{9}{*}{81{,}657}
& \textbf{Our Model} & \textbf{6.89} & \textbf{16.18\%} & \textcolor{black}{\textbf{11.10}}\\
&&& LightGBM & \textcolor{black}{7.16} & \textcolor{black}{17.62\%} & \textcolor{black}{11.82}\\
&&& CatBoost & \textcolor{black}{7.82} & \textcolor{black}{19.03\%} & \textcolor{black}{12.47}\\
&&& XGBoost & \textcolor{black}{7.80} & \textcolor{black}{20.22\%} & \textcolor{black}{12.25}\\
&&& Stacking & \textcolor{black}{7.19} & \textcolor{black}{18.38\%} & \textcolor{black}{11.31}\\
&&& LSTM & 7.21 & 17.08\% & \textcolor{black}{11.45}\\
&&& iTransformer & 7.45 & 18.04\% & \textcolor{black}{11.65}\\
&&& \textcolor{black}{Autoformer} & \textcolor{black}{7.31} & \textcolor{black}{18.04\%} & \textcolor{black}{11.73}\\
&&& \textcolor{black}{FEDformer} & \textcolor{black}{7.33} & \textcolor{black}{18.10\%} & \textcolor{black}{11.72}\\
\midrule
\multirow{9}{*}{72h--168h} & \multirow{9}{*}{186} & \multirow{9}{*}{28{,}460}
& \textbf{Our Model} & \textbf{11.97} & \textbf{11.50\%} & \textcolor{black}{\textbf{17.86}}\\
&&& LightGBM & \textcolor{black}{13.22} & \textcolor{black}{12.56\%} & \textcolor{black}{20.95}\\
&&& CatBoost & \textcolor{black}{15.88} & \textcolor{black}{14.80\%} & \textcolor{black}{24.22}\\
&&& XGBoost & \textcolor{black}{13.85} & \textcolor{black}{13.77\%} & \textcolor{black}{21.44}\\
&&& Stacking & \textcolor{black}{12.77} & \textcolor{black}{12.56\%} & \textcolor{black}{19.92}\\
&&& LSTM & 12.92 & 12.63\% & \textcolor{black}{18.47}\\
&&& iTransformer & 12.72 & 12.35\% & \textcolor{black}{18.21}\\
&&& \textcolor{black}{Autoformer} & \textcolor{black}{12.42} & \textcolor{black}{12.39\%} & \textcolor{black}{17.89}\\
&&& \textcolor{black}{FEDformer} & \textcolor{black}{12.42} & \textcolor{black}{12.43\%} & \textcolor{black}{17.90}\\
\midrule
\multirow{9}{*}{168h--504h} & \multirow{9}{*}{39} & \multirow{9}{*}{4{,}386}
& \textbf{Our Model} & \textbf{26.92} & \textbf{10.94\%} & \textcolor{black}{41.51}\\
&&& LightGBM & \textcolor{black}{29.37} & \textcolor{black}{12.20\%} & \textcolor{black}{47.02}\\
&&& CatBoost & \textcolor{black}{38.09} & \textcolor{black}{14.73\%} & \textcolor{black}{57.15}\\
&&& XGBoost & \textcolor{black}{30.47} & \textcolor{black}{13.19\%} & \textcolor{black}{45.47}\\
&&& Stacking & \textcolor{black}{27.05} & \textcolor{black}{11.72\%} & \textcolor{black}{\textbf{40.73}}\\
&&& LSTM & 27.77 & 11.55\% & \textcolor{black}{41.52}\\
&&& iTransformer & 28.86 & 12.45\% & \textcolor{black}{41.48}\\
&&& \textcolor{black}{Autoformer} & \textcolor{black}{28.51} & \textcolor{black}{68.55\%} & \textcolor{black}{44.04}\\
&&& \textcolor{black}{FEDformer} & \textcolor{black}{28.46} & \textcolor{black}{68.40\%} & \textcolor{black}{43.98}\\
\bottomrule
\end{tabularx}
    \begin{tablenotes}
        \item [a] Segment classification criteria: range requirements for average sailing durations in 2021.
        \item [b] The number of segments in the group.
        \item [c] The summation of the number of voyage records in 2021 for all segments in the group.
    \end{tablenotes}
\end{threeparttable}
\end{table}

Table~\ref{tab:freq_bins} summarizes the forecasting results after grouping segments by \emph{the number of sailing records} in 2021. The proposed model \textcolor{black}{achieves the best RMSE in all frequency groups}, and also attains the best or tied-best MAE in all groups. It further delivers the best MAPE in the three lower-frequency groups, while remaining competitive in the highest-frequency group. For segment group with \textcolor{black}{76}--150 records in 2021, it decreases MAE and MAPE by \textcolor{black}{\textbf{0.35} h} (\textcolor{black}{\textbf{4.53\%}} relatively) and \textbf{1.13} percentage points (\textbf{6.36\%} relatively), \textcolor{black}{also reducing RMSE by \textbf{0.24} h (\textbf{1.59\%} relatively)}. For segment group with 151--500 records, MAE is reduced by \textbf{0.28} h (\textbf{4.75\%} relatively), MAPE by \textbf{0.93} percentage points (\textbf{4.99\%} relatively) \textcolor{black}{and RMSE by \textbf{0.25} h (\textbf{2.33\%} relatively)}. For segment group with 501--1000 records, the proposed model achieves \textcolor{black}{the best MAE performance}, with \textbf{0.85} percentage points reduction in MAPE (\textbf{3.52\%} relatively) \textcolor{black}{and \textbf{0.02} h reduction in RMSE (\textbf{0.22\%} relatively)}. For the highest-frequency segment group (1001--5000 sailing records in 2021), the proposed model maintains comparable MAE and MAPE with \textcolor{black}{the sequential baseline models and lowest RMSE} (\textbf{2.91} h in MAE, \textbf{23.78\%} in MAPE, \textcolor{black}{and \textbf{6.01} h in RMSE}). Results for tree-based models lag markedly in every group. According to Figure~\ref{fig:segment} which shows medium-to-high-frequency segments correspond to medium-to-short sailing distances, the conclusion derived from Table~\ref{tab:freq_bins} can be consistent with that from Table~\ref{tab:all_bins}. The proposed model performs competitively on medium-frequency segments because the dense record enables the attention mechanism to capture temporal relationship. Besides, the outperformance on low-frequency segments indicates that the learned patterns from other segments may be transferred to low-frequency segments.

\begin{table}[h]
\centering
\caption{\textcolor{black}{Performance comparison of models for different segment groups binned by the number of sailing records.}}
\label{tab:freq_bins}
\renewcommand{\arraystretch}{1.12}
\begin{threeparttable}
\begin{tabularx}{\textwidth}{lllXlll}
\toprule
\textbf{\#Records}\tnote{a} & \textbf{\#Segments}\tnote{b} & \textbf{Sum}\tnote{c} & \textbf{Model} & \textbf{MAE (h)} & \textbf{MAPE} & \textcolor{black}{\textbf{RMSE (h)}}\\
\midrule
\multirow{9}{*}{\textcolor{black}{76}--150} & \multirow{9}{*}{\textcolor{black}{682}} & \multirow{9}{*}{70{,}362}
& \textbf{Our Model} & \textbf{7.37} & \textbf{16.63\%} & \textcolor{black}{\textbf{14.86}}\\
&&& LightGBM & \textcolor{black}{8.26} & \textcolor{black}{26.32\%} & \textcolor{black}{18.21}\\
&&& CatBoost & \textcolor{black}{9.94} & \textcolor{black}{38.12\%} & \textcolor{black}{20.24}\\
&&& XGBoost & \textcolor{black}{9.00} & \textcolor{black}{39.53\%} & \textcolor{black}{17.43}\\
&&& Stacking & \textcolor{black}{7.96} & \textcolor{black}{30.42\%} & \textcolor{black}{15.65}\\
&&& LSTM & 7.74 & 17.76\% & \textcolor{black}{15.10}\\
&&& iTransformer & 7.91 & 18.52\% & \textcolor{black}{15.15}\\
&&& \textcolor{black}{Autoformer} & \textcolor{black}{7.74} & \textcolor{black}{23.38\%} & \textcolor{black}{15.35}\\
&&& \textcolor{black}{FEDformer} & \textcolor{black}{7.72} & \textcolor{black}{23.29\%} & \textcolor{black}{15.26}\\
\midrule

\multirow{9}{*}{151--500} & \multirow{9}{*}{403} & \multirow{9}{*}{99{,}639}
& \textbf{Our Model} & \textbf{5.62} & \textbf{17.71\%} & \textcolor{black}{\textbf{10.46}}\\
&&& LightGBM & \textcolor{black}{5.92} & \textcolor{black}{31.22\%} & \textcolor{black}{11.03}\\
&&& CatBoost & \textcolor{black}{6.69} & \textcolor{black}{41.51\%} & \textcolor{black}{12.44}\\
&&& XGBoost & \textcolor{black}{6.70} & \textcolor{black}{52.76\%} & \textcolor{black}{11.90}\\
&&& Stacking & \textcolor{black}{5.94} & \textcolor{black}{37.16\%} & \textcolor{black}{10.71}\\
&&& LSTM & 5.90 & 18.64\% & \textcolor{black}{10.84}\\
&&& iTransformer & 6.06 & 20.85\% & \textcolor{black}{11.06}\\
&&& \textcolor{black}{Autoformer} & \textcolor{black}{5.94} & \textcolor{black}{23.94\%} & \textcolor{black}{10.86}\\
&&& \textcolor{black}{FEDformer} & \textcolor{black}{5.96} & \textcolor{black}{23.88\%} & \textcolor{black}{10.89}\\
\midrule

\multirow{9}{*}{501--1000} & \multirow{9}{*}{26} & \multirow{9}{*}{18{,}809}
& \textbf{Our Model} & \textbf{4.96} & \textbf{23.27\%} & \textcolor{black}{\textbf{9.23}}\\
&&& LightGBM & \textcolor{black}{5.41} & \textcolor{black}{35.87\%} & \textcolor{black}{10.60}\\
&&& CatBoost & \textcolor{black}{5.70} & \textcolor{black}{63.65\%} & \textcolor{black}{10.52}\\
&&& XGBoost & \textcolor{black}{6.16} & \textcolor{black}{125.95\%} & \textcolor{black}{10.98}\\
&&& Stacking & \textcolor{black}{5.52} & \textcolor{black}{68.55\%} & \textcolor{black}{10.16}\\
&&& LSTM & 5.21 & 24.12\% & \textcolor{black}{9.85}\\
&&& iTransformer & 5.11 & 27.10\% & \textcolor{black}{9.43}\\
&&& \textcolor{black}{Autoformer} & \textcolor{black}{\textbf{4.96}} & \textcolor{black}{25.36\%} & \textcolor{black}{9.31}\\
&&& \textcolor{black}{FEDformer} & \textcolor{black}{5.00} & \textcolor{black}{26.12\%} & \textcolor{black}{9.25}\\
\midrule

\multirow{9}{*}{1001--5000} & \multirow{9}{*}{9} & \multirow{9}{*}{12{,}700}
& \textbf{Our Model} & \textbf{2.91} & 23.78\% & \textcolor{black}{\textbf{6.01}}\\
&&& LightGBM & \textcolor{black}{3.37} & \textcolor{black}{31.72\%} & \textcolor{black}{7.07}\\
&&& CatBoost & \textcolor{black}{4.21} & \textcolor{black}{44.55\%} & \textcolor{black}{10.42}\\
&&& XGBoost & \textcolor{black}{4.58} & \textcolor{black}{63.40\%} & \textcolor{black}{8.67}\\
&&& Stacking & \textcolor{black}{3.51} & \textcolor{black}{35.70\%} & \textcolor{black}{7.09}\\
&&& LSTM & \textbf{2.91} & 23.52\% & \textcolor{black}{6.23}\\
&&& iTransformer & 3.09 & 24.91\% & \textcolor{black}{6.08}\\
&&& \textcolor{black}{Autoformer} & \textcolor{black}{2.93} & \textcolor{black}{23.12\%} & \textcolor{black}{6.19}\\
&&& \textcolor{black}{FEDformer} & \textcolor{black}{2.95} & \textcolor{black}{\textbf{22.28\%}} & \textcolor{black}{6.26}\\
\bottomrule
\end{tabularx}
    \begin{tablenotes}
        \item [a] Segment classification criteria: range requirements for the number of voyage records in 2021.
        \item [b] The number of segments in the group.
        \item [c] The summation of the number of voyage records in 2021 for all segments in the group.
    \end{tablenotes}
\end{threeparttable}
\end{table}

\begin{figure}[h]
    \centering
    \includegraphics[width=0.95\textwidth]{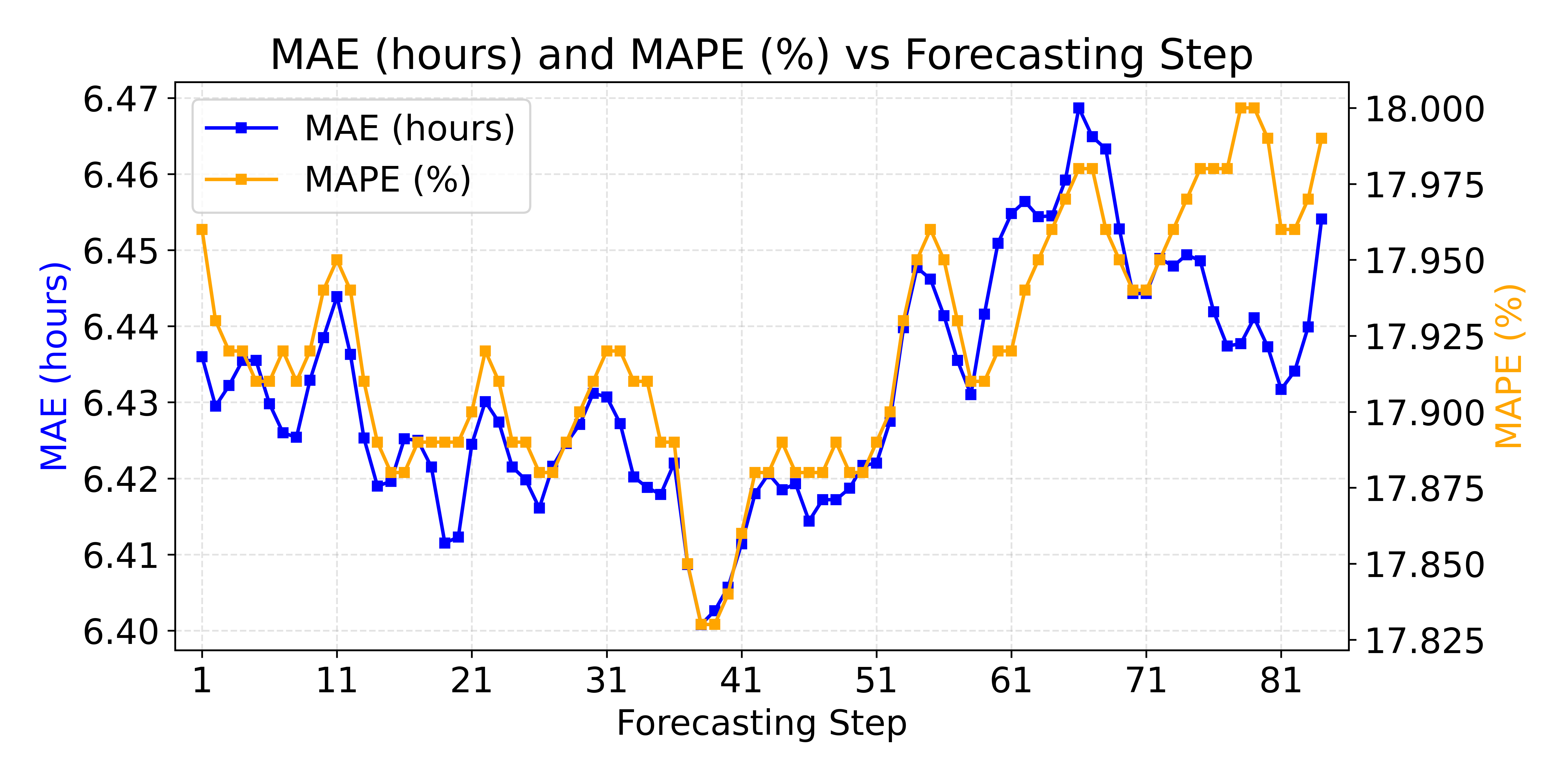}
    \caption{Prediction MAE and MAPE for different future time steps of the proposed model.}
    \label{fig:per_step}
\end{figure}

\begin{figure}[h]
    \centering
    \includegraphics[width=0.95\textwidth]{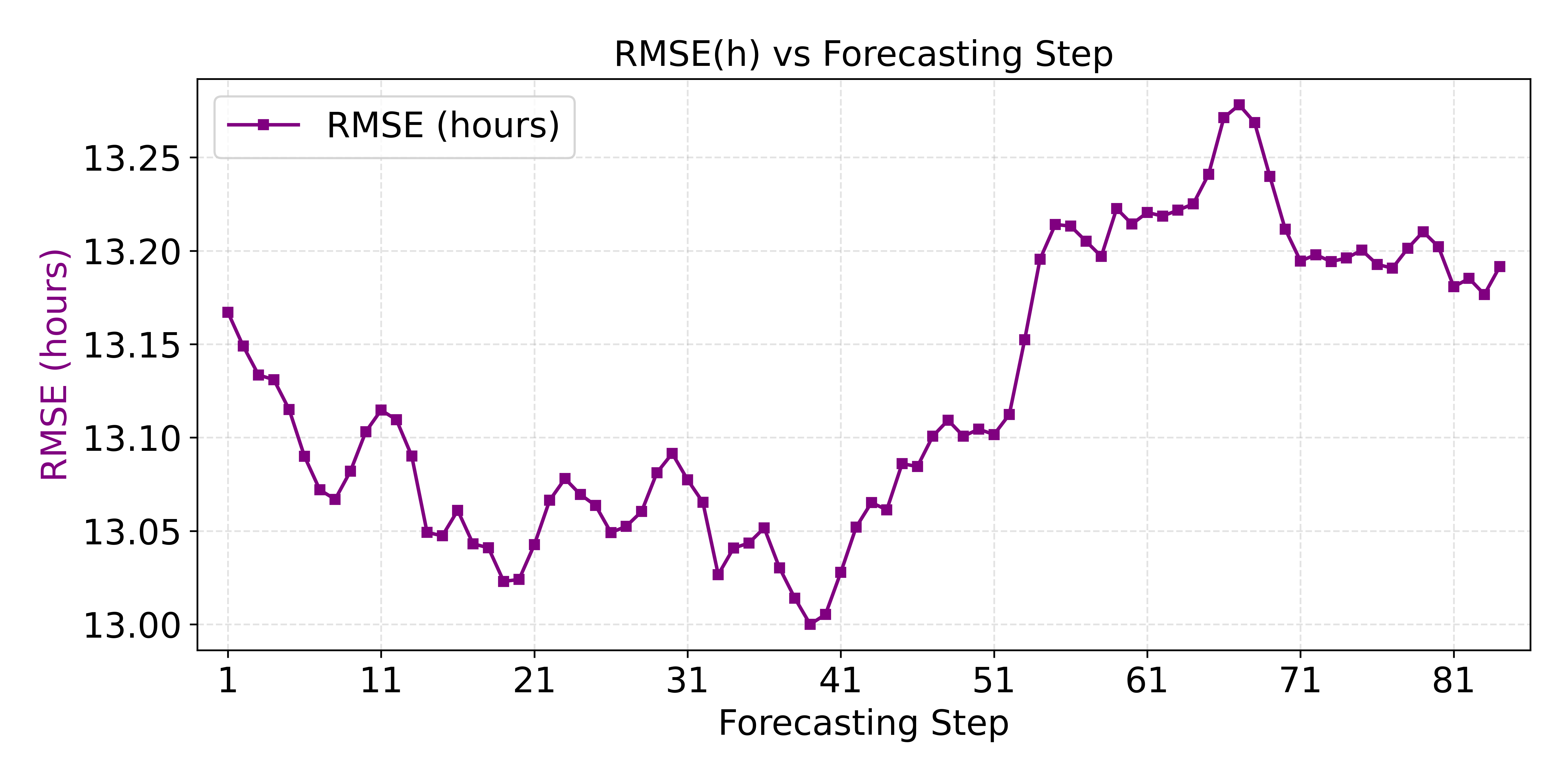}
    \caption{\textcolor{black}{Prediction RMSE for different future time steps of the proposed model.}}
    \label{fig:per_step_rmse}
\end{figure}

Figure~\ref{fig:per_step} \textcolor{black}{and Figure~\ref{fig:per_step_rmse}} show the performance metrics of the proposed model on different future time steps. Due to the sliding window mechanism shown in Figure~\ref{fig:Sliding_window}, not all observations have prediction results on all future time steps. For example, the first observation in the test set only has the prediction result on the first future time step and the last observation only has the prediction result on the last future time step. So, only observations with predictions on all future time steps are considered in this section to make sure the fairness of comparison, which causes a slight difference with overall results in performance metrics. The result indicates that both absolute and relative errors remain in a narrow band across the forecast horizon, suggesting stable long-term performance. The MAE fluctuates gently between \textbf{6.40}--\textbf{6.47}~h, while MAPE stays within \textbf{17.82\%}--\textbf{18.00\%} \textcolor{black}{and RMSE keeps between \textbf{13.00}--\textbf{13.30}~h.} No abrupt spikes or outliers appear in the horizon, showing that the model's error profile is smooth, bounded, and robust over time.

\subsection{Ablation Study}

To verify the effectiveness of input features and time series model design, 4 ablation experiments are conducted. The first one removes the time series features and the model almost falls back to simple MLP network. The second one removes features about vessels. The third one removes features related to the specific segment. The fourth one removes the auxiliary task. For each experiment, a new model is trained from scratch.


Table~\ref{tab:ablation} summarizes results of the ablation study after removing one feature group or auxiliary task at a time. The large degradations arise when time series features or segment features are removed. For the experiment removing time series features, MAE increases by \textbf{0.34} h, MAPE by \textbf{0.68} percentage points, \textcolor{black}{and RMSE by \textbf{2.17} h}, which hurts the absolute forecasting accuracy significantly. Eliminating segment features also leads to performance degradation, with MAE increasing by \textbf{0.17} h, MAPE increasing by \textbf{3.26} percentage points, \textcolor{black}{and RMSE increasing by \textbf{0.97} h}. This suggests that segment identifiers which record segment characteristics are particularly important for the relative accuracy of predictions. Removing vessel features leads to \textbf{0.18} h increase in MAE, \textbf{0.67} percentage points increase in MAPE, \textcolor{black}{and \textbf{1.02} h increase in RMSE}. If the auxiliary task is removed, MAE increases by \textbf{0.06} h, MAPE increases by \textbf{0.31} percentage points, \textcolor{black}{and RMSE increases by \textbf{0.07} h}, which indicates that the task of vessel number predictions in destination ports shares some coordinated signals with the task of sailing duration predictions. Overall, time series features, vessel features, segment features and the auxiliary task design all contribute to the improvement of prediction accuracy. Time series features and segment features contribute the most, followed by vessel characteristics and the auxiliary task.

\begin{table}[h]
\centering
\caption{\textcolor{black}{Ablation experiment results of the proposed model.}}
\label{tab:ablation}
\renewcommand{\arraystretch}{1.15}
\small
\begin{tabularx}{\textwidth}{
@{}l
>{\raggedright\arraybackslash}X
>{\centering\arraybackslash}p{0.145\textwidth}
>{\centering\arraybackslash}p{0.175\textwidth}
>{\centering\arraybackslash}p{0.145\textwidth}
@{}}
\toprule
\textbf{Setting} 
& \textbf{Removed features or task} 
& \makecell[c]{\textbf{MAE}\\\textbf{($\Delta$MAE, h)}} 
& \makecell[c]{\textbf{MAPE}\\\textbf{($\Delta$MAPE)}} 
& \makecell[c]{\textcolor{black}{\textbf{RMSE}}\\\textcolor{black}{\textbf{($\Delta$RMSE, h)}}} \\
\midrule
Full model 
& -- 
& \textbf{6.08} (--) 
& \textbf{18.05\%} (--) 
& \textcolor{black}{\textbf{12.04} (--)} \\

Ablation\,1 
& Time series features: sailing duration sequence, vessel count sequence 
& 6.42 (+0.34) 
& 18.73\% (+0.68 pp) 
& \textcolor{black}{14.21 (+2.17)} \\

Ablation\,2 
& Vessel features: width, length, TEU, carrier 
& 6.26 (+0.18) 
& 18.72\% (+0.67 pp) 
& \textcolor{black}{13.06 (+1.02)} \\

Ablation\,3 
& Segment features: start port, destination port, terminal 
& 6.25 (+0.17) 
& 21.31\% (+3.26 pp) 
& \textcolor{black}{13.01 (+0.97)} \\

Ablation\,4 
& Auxiliary task 
& 6.14 (+0.06) 
& 18.36\% (+0.31 pp) 
& \textcolor{black}{12.11 (+0.07)} \\
\bottomrule
\end{tabularx}
\end{table}

\subsection{\textcolor{black}{Inference Speed Testing}}

\textcolor{black}{To evaluate whether the model satisfies the low-latency requirement for practical deployment in shipping schedule applications, an inference speed experiment was conducted. Specifically, the recorded inference time corresponds to predicting the sailing durations of all selected routes over the next $H$ future time steps on the test dataset, where $H=84$. The inference time of the proposed model and all baseline models was measured and compared under the same experimental setting.}

\textcolor{black}{Table~\ref{tab:ref_time} reports the inference time of different models. The proposed model requires $0.883\ s$ to complete inference on the entire test dataset, which is slightly slower than the gradient boosting models. Overall, all models finish inference within approximately one second with only Autoformer slightly exceeding one second, indicating that the proposed model achieves competitive predictive performance without introducing prohibitive computational latency. This level of efficiency is sufficient to support practical deployment in shipping schedule applications where timely decision-making is required.}

\begin{table}[h]
\centering
\caption{\textcolor{black}{Inference time comparison for predicting sailing times over the next $H=84$ future time steps on the test dataset.}}
\label{tab:ref_time}
\renewcommand{\arraystretch}{1.12}
\begin{tabularx}{\textwidth}{X X}
\toprule
\textcolor{black}{\textbf{Model}} & \textcolor{black}{\textbf{Inference time(s)}} \\
\midrule
\textcolor{black}{Our Model} & \textcolor{black}{0.883}\\
\textcolor{black}{LightGBM} & \textcolor{black}{0.344}\\
\textcolor{black}{CatBoost} & \textcolor{black}{0.018}\\
\textcolor{black}{XGBoost} & \textcolor{black}{0.151}\\
\textcolor{black}{Stacking} & \textcolor{black}{0.541}\\
\textcolor{black}{LSTM} & \textcolor{black}{0.409}\\
\textcolor{black}{iTransformer} & \textcolor{black}{0.818}\\
\textcolor{black}{FEDformer} & \textcolor{black}{0.712}\\
\textcolor{black}{Autoformer} & \textcolor{black}{1.022}\\
\bottomrule
\end{tabularx}
\end{table}

\subsection{Sample Size Sensitivity Analysis}

To examine whether the predictive performance of the proposed model is significantly biased toward voyage segments with larger sample sizes,  a sensitivity analysis is conducted with respect to the number of observed sailing records per segment. However, since different segments exhibit very different distributions of sailing durations, raw MAE is not directly comparable across segments. To remove the influence of scale, both ground truth sailing durations and model predictions within each leg are standardized with Z-Score normalization. Afterwards, MAE with normalized target is computed as the segment prediction performance indicator.

\begin{figure}[h]
    \centering
    \includegraphics[width=0.95\textwidth]{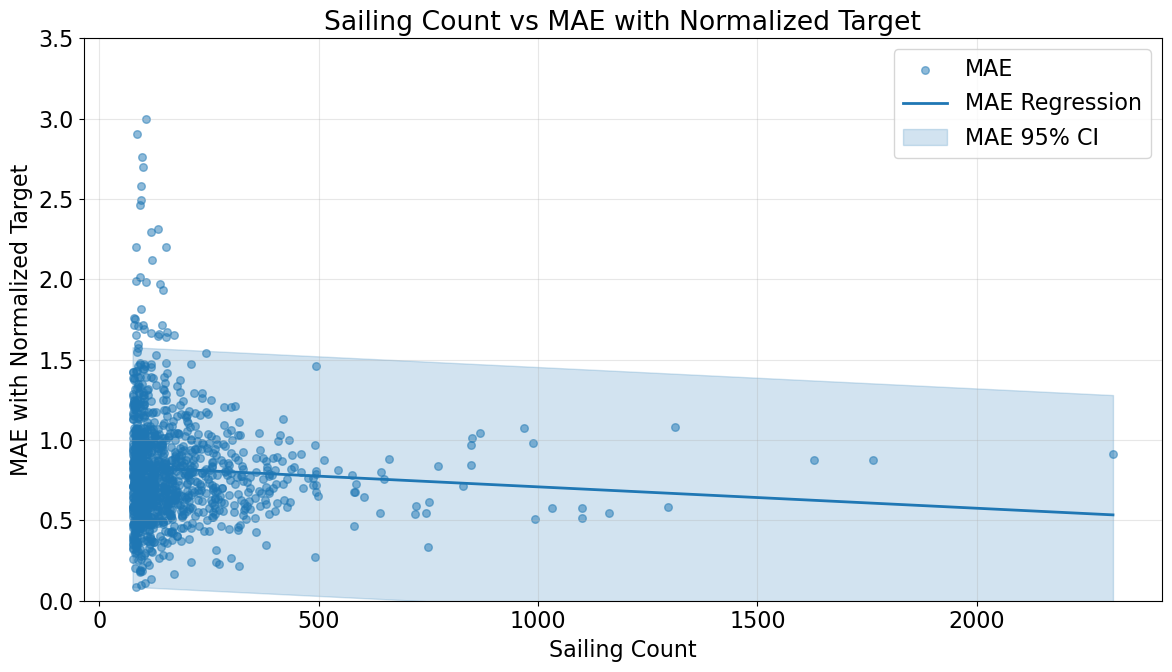}
    \caption{Scatter plot and linear regression line of MAE with normalized target with corresponding sailing counts.}
    \label{fig:sensitivity}
\end{figure}

Let $d^{(ij)}$ denote the number of sailing records observed for segment $\textcolor{black}{p^{(i)}} \rightarrow \textcolor{black}{p^{(j)}}$, and let $\mathrm{MAE}_{\mathrm{norm}}^{(ij)}$ be the corresponding normalized error metrics. The relationship between $d^{(ij)}$ and the indicator is investigated by fitting simple linear regression models:
\begin{equation*}
  \mathrm{MAE}_{\mathrm{norm}}^{(ij)}
  = \alpha_{\mathrm{MAE}} + \beta_{\mathrm{MAE}} d^{(ij)} + \varepsilon^{(ij)},
\end{equation*}
where $\alpha_{\mathrm{MAE}}$, $\beta_{\mathrm{MAE}}$ are learnable parameters and $\varepsilon^{(ij)}$ is the residual. Figure~\ref{fig:sensitivity} shows the relationship concretely.

The estimated slope for normalized MAE is $\beta_{\mathrm{MAE}} \approx -1.33\times 10^{-4}$ with a $95\%$ confidence interval (CI) $[-2.64\times 10^{-4}, -2\times 10^{-6}]$. Although a statistically significant negative trend is detected, its practical magnitude is minimal. Segments with many historical observations and legs with a limited number of observations exhibit comparable normalized MAE, and the variance is larger for segments with fewer sailing records. The prediction error remains roughly stable over a wide range of sailing counts, which provides evidence that the model maintains robust generalization capability across voyage legs with different frequency.

\subsection{Effect of Time Series Features}

\begin{figure}[h]
    \centering
    \includegraphics[width=0.95\textwidth]{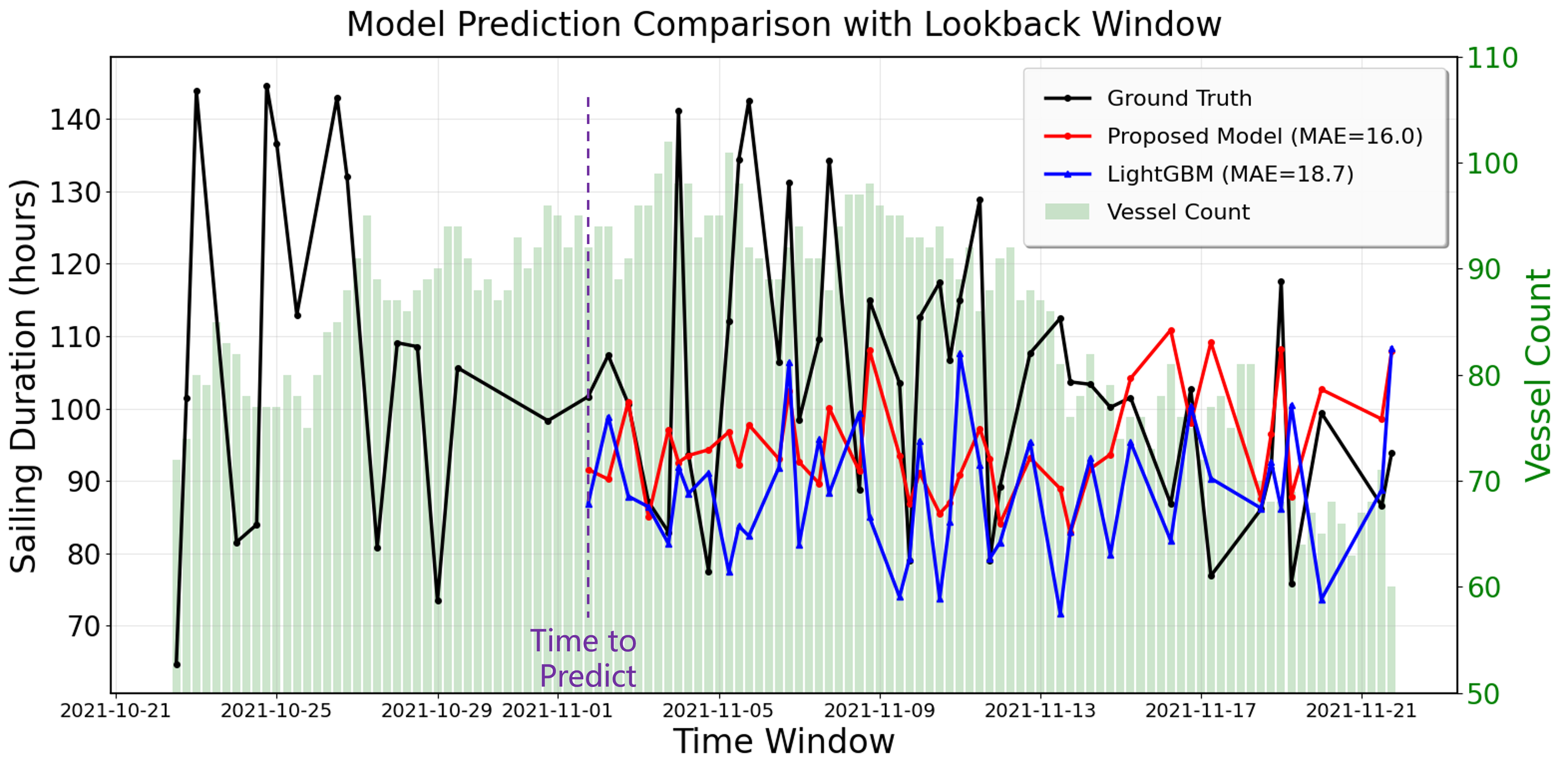}
    \caption{\textcolor{black}{Model prediction comparison for the proposed model and LightGBM (lookback window is truncated).}}
    \label{fig:prediction_comparison}
\end{figure}

Figure~\ref{fig:prediction_comparison} compares the prediction results of the proposed model and the LightGBM baseline on Shekou$\rightarrow$Singapore segment over a representative evaluation period. The proposed model, which incorporates historical sailing duration and port vessel count sequences, \textcolor{black}{achieves MAE of \textbf{15.97}~h, MAPE of \textbf{14.81\%} and RMSE of \textbf{20.12}~h, clearly outperforming the LightGBM baseline whose MAE, MAPE, RMSE are \textbf{18.71}~h, \textbf{16.78\%}, \textbf{24.32}~h respectively.}

This example illustrates the importance of time series features for accurate sailing duration forecasting on busy liner segments. In the training set, the average sailing duration on this leg is 90.7 hours, whereas during the plotted evaluation window the mean duration rises to 103.3 hours and the current vessel count is clearly high, reflecting the potential congestion around destination port in the future. By ingesting both historical sailing time sequences and the time series of vessel counts, the attention mechanism can perceive this upward shift in congestion and adjusts its predictions accordingly, with an average forecast of 94.87 hours that better tracks the increasing pattern. In contrast, LightGBM relies mainly on static features and lacks explicit temporal context, so its predictions remain close to the long run mean value (average \textcolor{black}{88.34} hours), systematically underestimating the actual sailing durations. The superior MAE and MAPE of the proposed model stem directly from its ability to exploit dynamic time series features to capture variations in port congestion.

\subsection{Interpretation of Attention Mechanism}

\begin{figure}[h]
    \centering
    \includegraphics[width=0.8\textwidth]{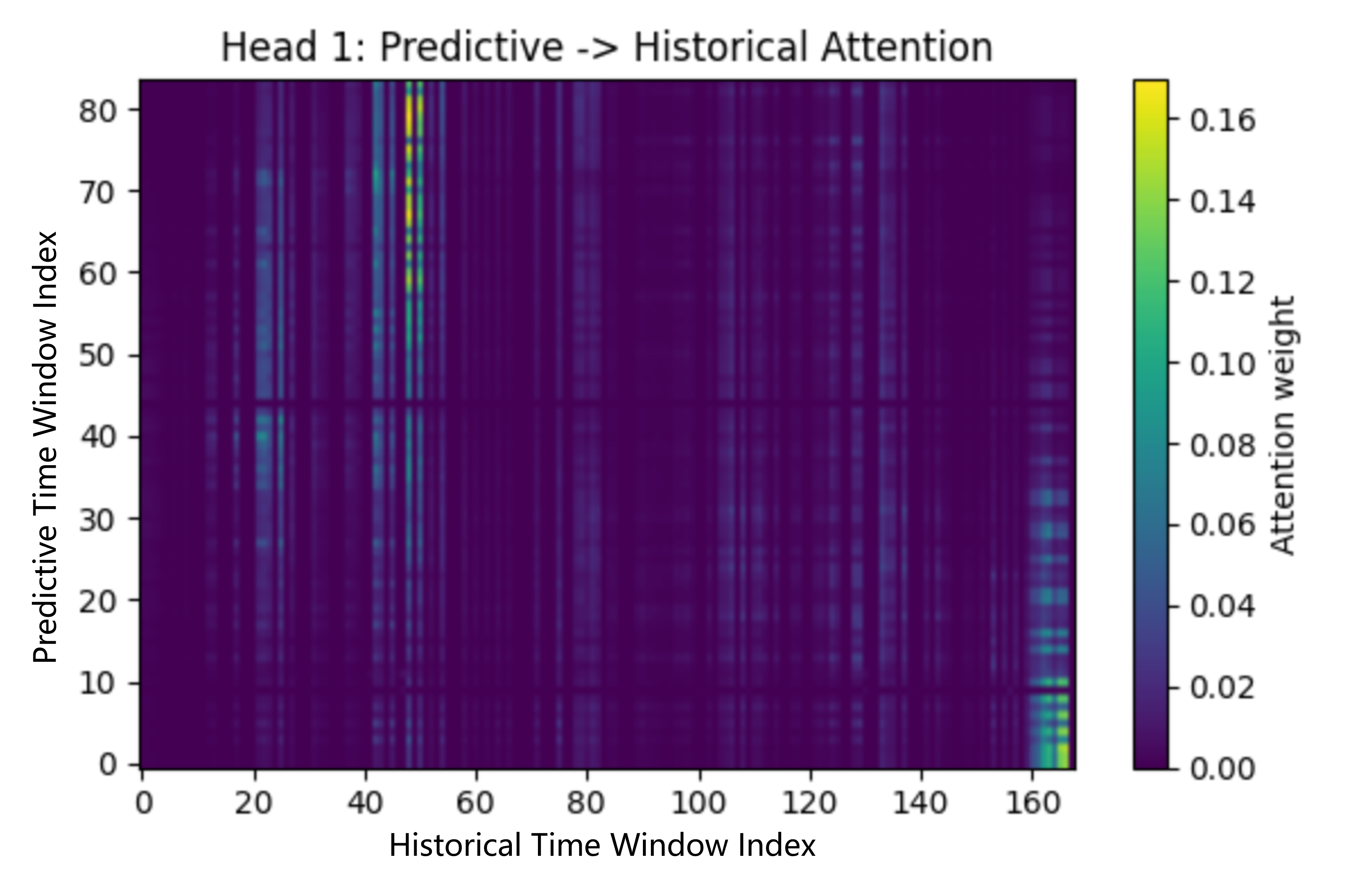}
    \caption{Attention weight heatmap between future and historical steps.}
    \label{fig:attn}
\end{figure}

Figure~\ref{fig:attn} visualizes the transformer attention from each predictive step (vertical axis) to historical steps (horizontal axis) for a sample on Shekou$\rightarrow$Singapore segment. The pattern is sparse and structured rather than average over the history, which means the attention mechanism \textcolor{black}{learns meaningful temporal dependencies}. Firstly, a persistent bright band appears from the near future to the most recent history (lower-right region), showing that short lag context remains the primary anchor for the prediction result in near future. Secondly, several vertical stripes at mid-range lags around indices 50-60 attract repeated focus, especially for later forecast steps, indicating that the model reuses specific historical regimes that resemble the current forecast context. Lastly, attention strength intensifies for a narrow cluster of historical positions while remaining low elsewhere, suggesting selective retrieval instead of uniform smoothing. The concentration of weights at a few salient lags helps stabilize long-horizon predictions by anchoring them to representative historical states rather than averaging over heterogeneous periods.

\subsection{\textcolor{black}{Sensitivity of Forecasting Horizon}}

The sensitivity of the proposed model is further investigated with respect to the forecasting horizon. The experiment is designed to evaluate the robustness of the model under different temporal prediction ranges. Concretely, the forecasting horizon $H$ is varied to evaluate the stability of prediction performance across different temporal ranges. Specifically, the forecasting horizon is set to 7 days ($H = 28$), 14 days ($H = 56$), and 21 days ($H = 84$), with all other hyperparameters kept unchanged. The result is reported in Table~\ref{tab:horizon_sensitivity}.


The model delivers consistently accurate predictions across all tested horizons, demonstrating its flexibility in capturing relevant patterns for forecasting horizons with different lengths. A detailed examination reveals an expected trend. As $H$ decreases, the predictive precision improves with MAE measuring \textbf{6.00} h at $H=28$ compared to \textbf{6.08} h at $H=84$, \textcolor{black}{and RMSE decreasing from \textbf{12.04} h to \textbf{11.89} h}. MAPE remains stable around \textbf{18\%}. This pattern aligns with the intuition that shorter forecasting horizons reduce future uncertainty, enabling more precise estimations. Overall, the results confirm that the model can effectively adapt to varying prediction lengths while maintaining strong performance. These findings indicate that the proposed architecture is not only capable of handling multi-horizon predictions but also responds appropriately to the varying uncertainty inherent in different forecast lengths.

\begin{table}[htbp]
\centering
\caption{\textcolor{black}{Sensitivity analysis with respect to forecasting horizon (model capacity fixed).}}
\label{tab:horizon_sensitivity}
\renewcommand{\arraystretch}{1.15}
\begin{tabularx}{\textwidth}{lXXX}
\toprule
\textbf{Forecasting Horizon} & \textbf{MAE (h)} & \textbf{MAPE} & \textcolor{black}{\textbf{RMSE (h)}}\\
\midrule
7 days ($H=28$)   & 6.00   & 17.82\% & \textcolor{black}{11.89}\\
14 days ($H=56$) & 6.06   & 18.09\% & \textcolor{black}{12.01}\\
21 days ($H=84$) & 6.08   & 18.05\% & \textcolor{black}{12.04}\\
\bottomrule
\end{tabularx}
\end{table}

\subsection{Case Study: Waigaoqiao Port}

\begin{figure}[htbp]
    \centering
    \includegraphics[width=0.95\textwidth]{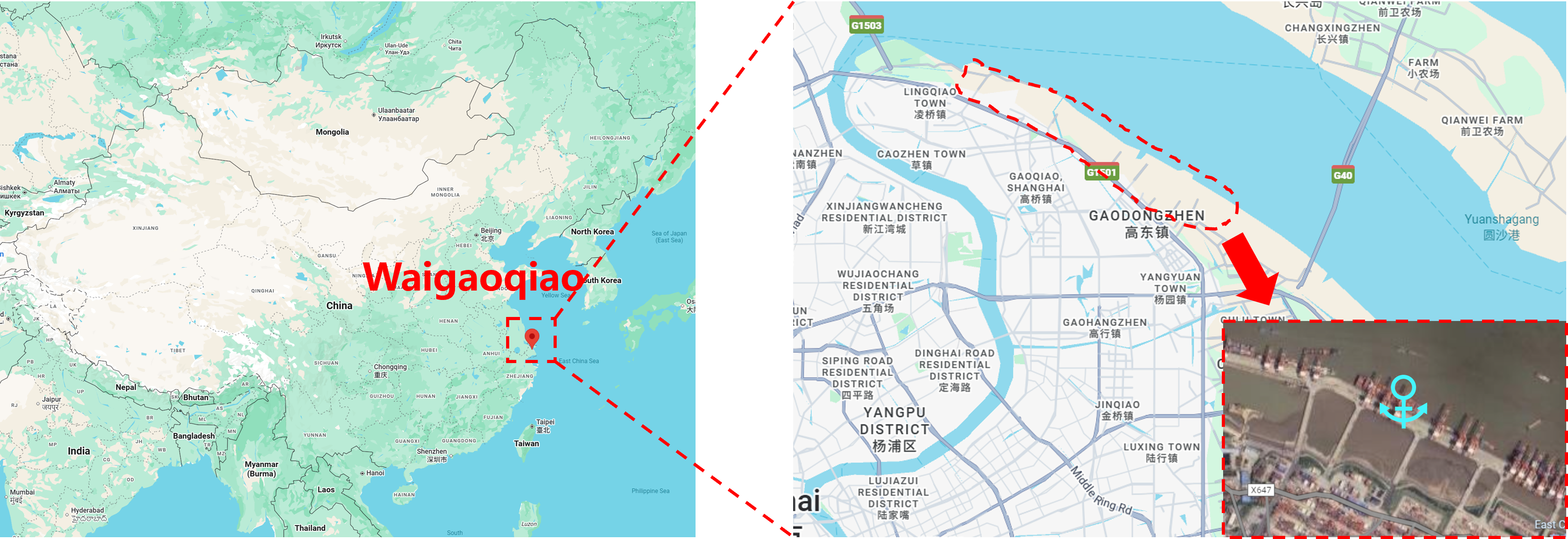}
    \caption{Waigaoqiao port location overview.}
    \label{fig:location}
\end{figure}

In this section, Waigaoqiao port is selected as the destination port for case study. Waigaoqiao (WGQ) is one of Shanghai's primary gateways linking the Yangtze River Delta to the global liner network. Figure~\ref{fig:location} shows the geographical depiction of WGQ. Although ultra-large mainline vessels increasingly berth at Yangshan Deep-Water Port, WGQ remains a critical node that stitches together East-West mainline routes and dense intra-Asia service loops. There are 25 segments in the study with WGQ as the destination. Table~\ref{tab:route_waigaoqiao} reports the prediction results on each segment for the proposed model and baseline model. The compared model is set to LSTM as it \textcolor{black}{shows competitive performances over other baselines}.

Based on the results, the proposed model attains lower MAE on \textbf{17 of 25} segments, lower MAPE on \textbf{16 of 25} segments \textcolor{black}{and lower RMSE on \textbf{17 of 25} segments}, evidencing both absolute and relative accuracy improvements. The advantages are most pronounced on medium-to-high-frequency service lines, where segment-level temporal regularities and port congestion signals are more informative.

\section{Conclusion}
\label{sec:conclusion}

Persistent uncertainty in vessel sailing durations remains a fundamental bottleneck that degrades maritime schedule reliability and operational efficiency. In particular, the unpredictability of future voyage segments undermines long-term operational planning for downstream ports, leading to inflated costs and the suboptimal allocation of berth, yard, and labor resources. While most existing literature remains confined to estimating the remaining duration of the current segment using real-time AIS, this study addresses the underexplored challenge of forecasting sailing durations for future segments where real-time AIS data is inherently unavailable. We reformulate this problem as a segment-level, time-series forecasting task and bridge the existing research gap through a unified modeling framework.

Specifically, the proposed solution encompasses a robust data preprocessing pipeline that transforms raw AIS, port geofence, and static vessel information into structured voyage records and port vessel count dataset. Building upon this, we introduce a transformer-based architecture featuring causal masking and early feature fusion, which is specifically designed to capture long-range temporal dependencies and historical regularities in the absence of real-time AIS data. To further enhance predictive stability, a multi-task learning strategy is integrated to jointly forecast segment sailing durations and destination port congestion. This approach leverages shared latent signals between port operations and sailing behavior, effectively mitigating the systemic uncertainty associated with future voyage legs.

\begin{table}[htbp]
\centering
\caption{\textcolor{black}{Segment-level performance comparison between the proposed model and LSTM for Waigaoqiao as the destination port.}}
\label{tab:route_waigaoqiao}
\renewcommand{\arraystretch}{1.12}
\small
\begin{threeparttable}
\begin{tabularx}{\textwidth}{l l lXXXXXX}
\toprule
\multirow{2}{*}{\textbf{Start}} & \multirow{2}{*}{\textbf{Destination}} & \multirow{2}{*}{\textbf{Count}\tnote{a}} &
\multicolumn{3}{c}{\textbf{Our model}} & \multicolumn{3}{c}{\textbf{LSTM}} \\
\cmidrule(lr){4-6}\cmidrule(lr){7-9}
 &  &  & \scriptsize\textbf{MAE (h)} & \scriptsize\textbf{MAPE} & \scriptsize\textcolor{black}{\textbf{RMSE(h)}} & \scriptsize\textbf{MAE (h)} & \scriptsize\textbf{MAPE} & \scriptsize\textcolor{black}{\textbf{RMSE(h)}} \\
\midrule
Bugo & Waigaoqiao & 10 & \textbf{11.33} & \textbf{10.83\%} & \textcolor{black}{\textbf{14.33}} & 14.23 & 13.77\% & \textcolor{black}{17.99} \\
Busan & Waigaoqiao & 47 & \textbf{5.41} & \textbf{15.66\%} & \textcolor{black}{\textbf{9.94}} & 6.21 & 18.27\% & \textcolor{black}{10.54} \\
Busan New Port & Waigaoqiao & 51 & 8.22 & 24.17\% & \textcolor{black}{12.96} & \textbf{8.11} & \textbf{23.84\%} & \textcolor{black}{\textbf{12.52}} \\
Cat Lai & Waigaoqiao & 39 & \textbf{12.15} & \textbf{9.60\%} & \textcolor{black}{\textbf{16.67}} & 14.52 & 11.40\% & \textcolor{black}{18.90} \\
Gwangyang & Waigaoqiao & 83 & \textbf{3.55} & \textbf{13.61\%} & \textcolor{black}{\textbf{5.29}} & 3.71 & 14.72\% & \textcolor{black}{5.30} \\
Haiphong & Waigaoqiao & 18 & \textbf{9.66} & \textbf{9.73\%} & \textcolor{black}{12.56} & 10.25 & 10.35\% & \textcolor{black}{\textbf{12.55}} \\
Hong Kong & Waigaoqiao & 59 & \textbf{7.12} & \textbf{11.94\%} & \textcolor{black}{10.15} & 7.15 & 11.98\% & \textcolor{black}{\textbf{10.07}} \\
Incheon & Waigaoqiao & 23 & \textbf{2.67} & \textbf{9.22\%} & \textcolor{black}{\textbf{3.55}} & 3.08 & 11.27\% & \textcolor{black}{3.68} \\
Kaohsiung & Waigaoqiao & 59 & 10.18 & 20.86\% & \textcolor{black}{15.81} & \textbf{9.57} & \textbf{20.54\%} & \textcolor{black}{\textbf{14.36}} \\
Keelung & Waigaoqiao & 26 & \textbf{4.82} & \textbf{16.40\%} & \textcolor{black}{\textbf{6.73}} & 5.24 & 17.06\% & \textcolor{black}{7.58} \\
Kobe & Waigaoqiao & 51 & \textbf{7.79} & \textbf{12.51\%} & \textcolor{black}{\textbf{13.53}} & 8.81 & 13.90\% & \textcolor{black}{14.68} \\
Laem Chabang & Waigaoqiao & 7 & 9.46 & 6.54\% & \textcolor{black}{10.96} & \textbf{7.48} & \textbf{5.16\%} & \textcolor{black}{\textbf{9.57}} \\
Moji & Waigaoqiao & 24 & \textbf{6.34} & \textbf{16.86\%} & \textcolor{black}{\textbf{9.07}} & 7.43 & 20.84\% & \textcolor{black}{9.28} \\
Nagoya & Waigaoqiao & 35 & \textbf{5.61} & \textbf{9.22\%} & \textcolor{black}{9.28} & 6.04 & 10.40\% & \textcolor{black}{\textbf{8.98}} \\
Ningbo & Waigaoqiao & 137 & \textbf{1.92} & \textbf{32.64\%} & \textcolor{black}{\textbf{2.74}} & 2.05 & 35.34\% & \textcolor{black}{2.88} \\
Port Klang & Waigaoqiao & 16 & \textbf{13.46} & \textbf{9.16\%} & \textcolor{black}{\textbf{15.08}} & 15.41 & 9.87\% & \textcolor{black}{20.02} \\
Qingdao & Waigaoqiao & 106 & 5.18 & 20.07\% & \textcolor{black}{7.62} & \textbf{4.67} & \textbf{18.73\%} & \textcolor{black}{\textbf{6.89}} \\
Shimizu & Waigaoqiao & 19 & \textbf{4.34} & \textbf{7.66\%} & \textcolor{black}{\textbf{4.90}} & 4.44 & 7.86\% & \textcolor{black}{5.18} \\
Singapore & Waigaoqiao & 27 & 14.22 & 10.24\% & \textcolor{black}{\textbf{17.69}} & \textbf{12.79} & \textbf{8.82\%} & \textcolor{black}{20.68} \\
Taichung & Waigaoqiao & 16 & 5.13 & 14.59\% & \textcolor{black}{6.56} & \textbf{4.23} & \textbf{12.51\%} & \textcolor{black}{\textbf{5.26}} \\
Tokuyama & Waigaoqiao & 14 & \textbf{3.03} & \textbf{10.37\%} & \textcolor{black}{\textbf{3.37}} & 4.24 & 14.09\% & \textcolor{black}{4.82} \\
Tokyo & Waigaoqiao & 37 & 5.78 & 9.93\% & \textcolor{black}{\textbf{8.23}} & \textbf{5.73} & \textbf{9.66\%} & \textcolor{black}{8.28} \\
Xiamen & Waigaoqiao & 25 & \textbf{6.22} & \textbf{16.07\%} & \textcolor{black}{\textbf{7.96}} & 8.01 & 19.50\% & \textcolor{black}{9.73} \\
Yokohama & Waigaoqiao & 51 & 6.92 & 10.23\% & \textcolor{black}{\textbf{10.31}} & \textbf{6.71} & \textbf{9.79\%} & \textcolor{black}{10.58} \\
Zhoushan & Waigaoqiao & 25 & \textbf{4.38} & 27.98\% & \textcolor{black}{\textbf{7.95}} & 4.42 & \textbf{27.60\%} & \textcolor{black}{8.06} \\
\bottomrule
\end{tabularx}
\begin{tablenotes}
    \item [a] The number of voyage records on the specific segment in test dataset.
\end{tablenotes}
\end{threeparttable}
\end{table}

Empirically, the proposed model consistently outperforms competitive baselines across multiple evaluation protocols. Under record-weighted evaluation, it achieves MAE of \textbf{6.08} h, MAPE of \textbf{18.05\%} \textcolor{black}{and RMSE of \textbf{12.04 h}}, improving over the \textcolor{black}{competitive sequential deep learning models 4.70\% in MAE, 4.95\% in MAPE and 2.59\% in RMSE} relatively. Under the unweighted protocol, it reaches MAE of \textbf{7.23} h, MAPE of \textbf{17.21\%}, \textcolor{black}{and RMSE of \textbf{15.60 h}}, with respective gains of \textcolor{black}{4.37\%}, 5.65\% and \textcolor{black}{0.64\%} over baseline models relatively. The technical advantage over \textcolor{black}{the sequential baseline deep learning models (LSTM, iTransformer, FEDformer, Autoformer)} is primarily attributed to the early fusion of heterogeneous features before decoding and temporal attention mechanism, which enables richer cross-feature interactions aligned to each time window and allows the causal decoder to attend over a joint representation. This makes the proposed model outperform \textcolor{black}{baseline models} consistently and yield lower errors especially on medium-sailing-duration segments and medium-frequency segments. Furthermore, the model demonstrates a clear superiority over tree-based methods (e.g., LightGBM) by explicitly modeling segment-level historical temporal dynamics that flattened tabular snapshots cannot capture. These design choices result in improved accuracy for \textcolor{black}{59.19\%} of segments in terms of MAE, \textcolor{black}{62.77\%} in terms of MAPE and \textcolor{black}{60.54\% of segments in terms of RMSE}. Overall, this study provides a reliable decision-support framework that facilitates long-horizon planning and protects the integrity of transshipment chains across global liner services.

While this study establishes a robust framework for segment-level sailing duration prediction, several avenues remain for future exploration to enhance model reliability and operational depth. First, future research could focus on uncertainty quantification. Implementing probabilistic forecasting techniques such as quantile regression or diffusion-based models would produce calibrated interval predictions, allowing carriers to better manage planning risks \citep{Lee2021DataDrivenAF, wenzhe2025stgdpmvesseltrajectorypredictionspatiotemporal}. This could be further enriched by integrating dynamic environmental variables, including meteorological and oceanographic data, to enhance robustness under extreme conditions.

Second, the current framework could be extended to capture spatiotemporal dependencies across the global maritime network. Utilizing graph neural networks (GNN) over port-segment graphs would enable the modeling of systemic congestion propagation and inter-route dependencies. \textcolor{black}{In this way, the present segment-level framework can be extended to end-to-end multi-port ETA estimation. This would allow the cumulative impact of upstream delays to be explicitly modeled through sequential propagation. With the adjacent port congestion state and graph modeling, the model can produce more reliable prediction results.} Furthermore, incorporating latent route-choice modules could help account for unobserved path heterogeneity across different carrier preferences. \textcolor{black}{Future research may further explicitly characterize route-specific weather effects to enhance the prediction accuracy and reliability.} Finally, improving model generalization and operational utility is critical. Leveraging transfer learning or continual learning techniques would better support low-frequency segments with sparse historical data \citep{miao2024unifiedreplaybasedcontinuouslearning, yuan2024spatiotemporal}. Expanding the multi-task framework to predict additional port-level operations, such as turnaround time and queue length, would also provide a more holistic decision-support tool for synchronized liner service coordination.

\section*{Acknowledgments}

The acknowledgements section will be completed after the peer-review process.


\section*{Declaration of generative AI and AI-assisted technologies in the manuscript preparation process}

During the preparation of this work the authors used ChatGPT 5.2 in order to improve the language and assist with \LaTeX writing. After using this tool, the authors reviewed and edited the content as needed and take full responsibility for the content of the published article.

%
\begin{APPENDIX}{ }
\section{Raw Dataset Description}
\label{sec:raw}

AIS data, port geofence data and static vessel information serve as the raw datasets in Section~\ref{sec:process}. 

Automatic identification system (AIS) data refers to the dynamic and static information of vessels obtained through the shipborne AIS devices. The system periodically broadcasts information such as vessel name, position, speed, course, and destination port, while simultaneously receiving similar information from surrounding vessels. AIS data records the real-time position of vessels which is used for vessel tracking. Typical AIS data fields and their meanings are shown in Table~\ref{tab:ais_features}.

\begin{table}[h]
\centering
\caption{AIS data fields and descriptions}
\label{tab:ais_features}
\renewcommand{\arraystretch}{1.12}
\begin{tabularx}{1.0\textwidth}{XX}
\toprule
\textbf{Column name} & \textbf{Description} \\
\midrule
createTime & Timestamp of the AIS record \\ 
IMO & International Maritime Organization number \\ 
MMSI & Maritime Mobile Service Identity \\ 
speed & Vessel sailing speed (knots) \\ 
latitude & Latitude of the vessel's position \\ 
longitude & Longitude of the vessel's position \\ 
head & Vessel heading direction \\ 
draught & Vessel draught depth \\ 
destination & Destination port name \\ 
ETA & Estimated Time of Arrival reported by captain \\
\bottomrule
\end{tabularx}
\end{table}

Port geofence data records important spatial attributes of ports, including port locations, terminals, anchorage zones, pilot zones and berth boundaries. The data fields are summarized in Table~\ref{tab:port_geofence}. The anchorage, pilot and berth boundaries can be combined with vessel position from raw AIS data to determine the vessel's current behavior and operational status.

\begin{table}[h]
\centering
\caption{Port geofence data fields and descriptions}
\label{tab:port_geofence}
\renewcommand{\arraystretch}{1.12}
\begin{tabularx}{1.0\textwidth}{XX}
\toprule
\textbf{Column name} & \textbf{Description} \\
\midrule
portName & Port name \\ 
portId & Unique port identifier \\ 
tnmList & List of terminal identifiers \\ 
prkList & List of berth identifiers \\ 
waitingArea & Anchorage area boundary \\ 
pilotArea & Pilotage area boundary \\ 
parkingArea & Berth area boundary \\
\bottomrule
\end{tabularx}
\end{table}

Static vessel information consists of the physical and operational characteristics of \textcolor{black}{vessels}, such as vessel length, width, carrier, and maximum capacity. The main fields are summarized in Table~\ref{tab:vessel_static}. By combining static vessel information with raw AIS data, it becomes possible to analyze the navigation behavior and preferences of different types of vessels.

\begin{table}[htbp]
\centering
\caption{Vessel static feature data fields and their descriptions}
\label{tab:vessel_static}
\renewcommand{\arraystretch}{1.12}
\begin{tabularx}{1.0\textwidth}{XX}
\toprule
\textbf{Column name} & \textbf{Description} \\
\midrule
IMO & International Maritime Organization number\\ 
MMSI & Maritime Mobile Service Identity \\ 
crrName & Vessel carrier company \\ 
TEU & Maximum container capacity\\ 
width & Vessel width\\ 
length & Vessel length \\
\bottomrule
\end{tabularx}
\end{table}

\section{\textcolor{black}{Analysis of Port Congestion Indicators}}
\label{sec:congestion_indicator}

\textcolor{black}{To further examine the effectiveness of different port congestion indicators, we compare the port vessel count with two alternative congestion proxies: \textit{Detailed Vessel Count} and \textit{Anchorage Waiting Time}. The \textit{Detailed Vessel Count} proxy refines the original vessel count indicator by grouping vessels according to TEU capacity (0--3000 TEU, 3000--6000 TEU, 6000--12000 TEU, \textgreater 12000 TEU), thereby considering the heterogeneous occupation of port resources by vessels of different sizes. The \textit{Anchorage Waiting Time} proxy uses anchorage waiting time as a more direct indicator of port congestion. As shown in Table~\ref{tab:congestion_indicator}, the original \textit{Vessel Count} and \textit{Detailed Vessel Count} achieve very similar predictive performance, with average MAE values of 6.10 h and 6.11 h, respectively, while \textit{Anchorage Waiting Time} obtains a slightly higher average MAE of 6.15 h. Results suggest that two alternative congestion proxies do not provide a clear predictive advantage over the original port vessel count indicator in our experimental setting. For \textit{Detailed Vessel Count}, subdividing vessels by TEU increases the dimensionality of the input sequence, while some vessel TEU categories occur infrequently, leading to sparse and potentially noisy feature sequences. This may weaken the model's ability to learn stable congestion patterns from the refined features. For \textit{Anchorage Waiting Time}, although anchorage waiting time is closely related to congestion, it varies substantially across vessels and may exhibit strong volatility. Moreover, anchorage waiting time can reflect congestion with a certain time lag, as it is often observed after congestion has already affected vessel operations. In contrast, the total number of vessels in port provides a more stable and timely representation of the current level of port resource occupation. Therefore, the port vessel count is selected as the port congestion indicator in the model.}

\begin{table}[!htbp]
\centering
\caption{\textcolor{black}{Performance of different congestion proxies.}}
\label{tab:congestion_indicator}
\renewcommand{\arraystretch}{1.12}
\begin{tabularx}{\textwidth}{llXXXXX}
\toprule
\textbf{\textcolor{black}{Proxy}} & \textbf{\textcolor{black}{Metrics}} & \textbf{\textcolor{black}{Run 1}} & \textbf{\textcolor{black}{Run 2}} & \textbf{\textcolor{black}{Run 3}} & \textbf{\textcolor{black}{Mean}} & \textbf{\textcolor{black}{Std.}} \\
\midrule

\multirow{3}{*}{\textcolor{black}{Vessel Count}}
& \textcolor{black}{MAE (h)}  & \textcolor{black}{6.08} & \textcolor{black}{6.12} & \textcolor{black}{6.09} & \textcolor{black}{6.10} & \textcolor{black}{0.02} \\
& \textcolor{black}{MAPE}     & \textcolor{black}{18.05\%} & \textcolor{black}{18.04\%} & \textcolor{black}{18.05\%} & \textcolor{black}{18.05\%} & \textcolor{black}{0.01\%} \\
& \textcolor{black}{RMSE (h)} & \textcolor{black}{12.04} & \textcolor{black}{12.17} & \textcolor{black}{12.05} & \textcolor{black}{12.09} & \textcolor{black}{0.07} \\
\midrule

\multirow{3}{*}{\textcolor{black}{Detailed Vessel Count}}
& \textcolor{black}{MAE (h)}  & \textcolor{black}{6.09} & \textcolor{black}{6.11} & \textcolor{black}{6.13} & \textcolor{black}{6.11} & \textcolor{black}{0.02} \\
& \textcolor{black}{MAPE}     & \textcolor{black}{18.08\%} & \textcolor{black}{18.10\%} & \textcolor{black}{18.13\%} & \textcolor{black}{18.10\%} & \textcolor{black}{0.03\%} \\
& \textcolor{black}{RMSE (h)} & \textcolor{black}{12.26} & \textcolor{black}{12.08} & \textcolor{black}{12.10} & \textcolor{black}{12.15} & \textcolor{black}{0.10} \\
\midrule

\multirow{3}{*}{\textcolor{black}{Anchorage Waiting Time}}
& \textcolor{black}{MAE (h)}  & \textcolor{black}{6.14} & \textcolor{black}{6.14} & \textcolor{black}{6.17} & \textcolor{black}{6.15} & \textcolor{black}{0.02} \\
& \textcolor{black}{MAPE}     & \textcolor{black}{17.94\%} & \textcolor{black}{18.05\%} & \textcolor{black}{18.20\%} & \textcolor{black}{18.06\%} & \textcolor{black}{0.13\%} \\
& \textcolor{black}{RMSE (h)} & \textcolor{black}{12.03} & \textcolor{black}{12.07} & \textcolor{black}{12.16} & \textcolor{black}{12.09} & \textcolor{black}{0.07} \\

\bottomrule
\end{tabularx}
\end{table}

\section{Baseline Model Hyperparameter Settings}
\label{sec:hp}

This appendix summarizes the key hyperparameters used for all baseline models evaluated in this study. The configurations are reported to ensure reproducibility and to provide transparency regarding model capacity and regularization choices. All hyperparameters were fixed across experiments unless otherwise stated. The details are shown in Tables~\ref{tab:lgb_params}--\ref{tab:fed_params}.

\begin{table}[htbp]
\centering
\caption{\textcolor{black}{Hyperparameter configuration of the LightGBM baseline.}}
\renewcommand{\arraystretch}{1.12}
\label{tab:lgb_params}
\begin{tabularx}{1.0\textwidth}{XX}
\toprule
\textbf{Hyperparameter} & \textbf{Value} \\
\midrule
Objective function & MAE \\
Evaluation metric & MAE \\
Number of boosting rounds & \textcolor{black}{2186} \\
Learning rate & \textcolor{black}{0.0493} \\
Maximum tree depth & \textcolor{black}{20} \\
Number of leaves & \textcolor{black}{12483} \\
L1 regularization ($\lambda_{1}$) & \textcolor{black}{0.914} \\
L2 regularization ($\lambda_{2}$) & \textcolor{black}{0.0536} \\
\bottomrule
\end{tabularx}
\end{table}

\begin{table}[htbp]
\centering
\caption{\textcolor{black}{Hyperparameter configuration of the XGBoost baseline.}}
\label{tab:xgb_params}
\renewcommand{\arraystretch}{1.12}
\begin{tabularx}{1.0\textwidth}{XX}
\toprule
\textbf{Hyperparameter} & \textbf{Value} \\
\midrule
Objective function & Squared error regression \\
Number of estimators & \textcolor{black}{2110} \\
Learning rate & \textcolor{black}{0.0340} \\
Maximum tree depth & \textcolor{black}{12} \\
Min child weight & \textcolor{black}{13.420} \\
L1 regularization ($\alpha$) & \textcolor{black}{0.00488} \\
L2 regularization ($\lambda$) & \textcolor{black}{9.342} \\
Subsample & \textcolor{black}{0.955} \\
Fraction of features & \textcolor{black}{0.703} \\
Tree construction method & Histogram-based \\
\bottomrule
\end{tabularx}
\end{table}

\begin{table}[htbp]
\centering
\caption{\textcolor{black}{Hyperparameter configuration of the CatBoost baseline.}}
\label{tab:cat_params}
\renewcommand{\arraystretch}{1.12}
\begin{tabularx}{1.0\textwidth}{XX}
\toprule
\textbf{Hyperparameter} & \textbf{Value} \\
\midrule
Loss function & MAE \\
Number of trees & \textcolor{black}{1954} \\
Tree depth & \textcolor{black}{10} \\
Learning rate & \textcolor{black}{0.0960} \\
L2 leaf regularization & \textcolor{black}{19.923} \\
Random strength & \textcolor{black}{0.0345} \\
Bagging temperature & \textcolor{black}{6.809} \\
\bottomrule
\end{tabularx}
\end{table}

\begin{table}[htbp]
\centering
\caption{Configuration of the stacking ensemble baseline.}
\label{tab:stacking_params}
\renewcommand{\arraystretch}{1.12}
\begin{tabularx}{1.0\textwidth}{XX}
\toprule
\textbf{Component} & \textbf{Description} \\
\midrule
Base learners & LightGBM, XGBoost, CatBoost \\
Base learner hyperparameters & Identical to Tables~\ref{tab:lgb_params}--\ref{tab:cat_params} \\
Meta-learner & Linear regression \\
Training strategy & Out-of-fold prediction on training set \\
\bottomrule
\end{tabularx}
\end{table}

\begin{table}[htbp]
\centering
\caption{Hyperparameter configuration of the LSTM baseline.}
\label{tab:lstm_params}
\renewcommand{\arraystretch}{1.12}
\begin{tabularx}{1.0\textwidth}{XX}
\toprule
\textbf{Hyperparameter} & \textbf{Value} \\
\midrule
Hidden dimension & 64 \\
Number of LSTM layers & 2 \\
Embedding dimension & 16 \\
Dropout rate & 0.1 \\
\bottomrule
\end{tabularx}
\end{table}

\begin{table}[htbp]
\centering
\caption{Hyperparameter configuration of the iTransformer baseline.}
\label{tab:itransformer_params}
\renewcommand{\arraystretch}{1.12}
\begin{tabularx}{1.0\textwidth}{XX}
\toprule
\textbf{Hyperparameter} & \textbf{Value} \\
\midrule
Model dimension & 64 \\
Embedding dimension & 8 \\
Number of transformer layers & 1 \\
Number of attention heads & 8 \\
Feedforward dropout & 0.1 \\
Attention dropout & 0.1 \\
\bottomrule
\end{tabularx}
\end{table}

\begin{table}[htbp]
\centering
\caption{\textcolor{black}{Hyperparameter configuration of the Autoformer baseline.}}
\label{tab:auto_params}
\renewcommand{\arraystretch}{1.12}
\begin{tabularx}{1.0\textwidth}{XX}
\toprule
\textcolor{black}{\textbf{Hyperparameter}} & \textcolor{black}{\textbf{Value}} \\
\midrule
\textcolor{black}{Model dimension} & \textcolor{black}{16} \\
\textcolor{black}{Number of layers} & \textcolor{black}{2} \\
\textcolor{black}{Moving average window size} & \textcolor{black}{3} \\
\textcolor{black}{Feedforward dimension} & \textcolor{black}{64} \\
\bottomrule
\end{tabularx}
\end{table}

\begin{table}[htbp]
\centering
\caption{\textcolor{black}{Hyperparameter configuration of the FEDformer baseline.}}
\label{tab:fed_params}
\renewcommand{\arraystretch}{1.12}
\begin{tabularx}{1.0\textwidth}{XX}
\toprule
\textcolor{black}{\textbf{Hyperparameter}} & \textcolor{black}{\textbf{Value}} \\
\midrule
\textcolor{black}{Model dimension} & \textcolor{black}{16} \\
\textcolor{black}{Number of transformer layers} & \textcolor{black}{3} \\
\textcolor{black}{Version} & \textcolor{black}{Fourier} \\
\textcolor{black}{Modes} & \textcolor{black}{32} \\
\textcolor{black}{Mode selection} & \textcolor{black}{low} \\
\bottomrule
\end{tabularx}
\end{table}

\section{\textcolor{black}{Repeated Experiments with Different Random Seeds}}
\label{sec:repeat_experiments}

\textcolor{black}{To verify the repeatability and robustness of the proposed framework, we repeat the experiment three times using different random seeds while keeping the data split, preprocessing procedure, and selected hyperparameter settings unchanged. The results are reported in Table~\ref{tab:repeat_seed}. The small standard deviations across different runs indicate that the proposed model achieves stable and consistent predictive performance.}

\begin{table}[!htbp]
\centering
\caption{\textcolor{black}{Results of repeated experiments with different random seeds.}}
\label{tab:repeat_seed}
\begin{tabularx}{\textwidth}{XXXX}
\toprule
\textcolor{black}{\textbf{Run}} & \textcolor{black}{\textbf{MAE (h)}} & \textcolor{black}{\textbf{MAPE}} & \textcolor{black}{\textbf{RMSE (h)}} \\
\midrule
\textcolor{black}{Run 1} & \textcolor{black}{6.08} & \textcolor{black}{18.05\%} & \textcolor{black}{12.04} \\
\textcolor{black}{Run 2} & \textcolor{black}{6.12} & \textcolor{black}{18.04\%} & \textcolor{black}{12.17} \\
\textcolor{black}{Run 3} & \textcolor{black}{6.09} & \textcolor{black}{18.05\%} & \textcolor{black}{12.05} \\
\midrule
\textcolor{black}{Average} & \textcolor{black}{6.097} & \textcolor{black}{18.047\%} & \textcolor{black}{12.087} \\
\textcolor{black}{Standard deviation} & \textcolor{black}{0.021} & \textcolor{black}{0.006\%} & \textcolor{black}{0.073} \\
\bottomrule
\end{tabularx}
\end{table}

\end{APPENDIX}
%
%


\bibliographystyle{elsarticle-harv} 
\bibliography{sample} 

@inproceedings{10.1109/ITSC48978.2021.9564883,
author = {Noman, Abdullah A. and Heuermann, Aaron and Wiesner, Stefan A. and Thoben, Klaus-Dieter},
title = {Towards Data-Driven GRU based ETA Prediction Approach for Vessels on both Inland Natural and Artificial Waterways},
year = {2021},
publisher = {IEEE Press},
booktitle = {2021 IEEE International Intelligent Transportation Systems Conference (ITSC)},
pages = {2286–2291},
numpages = {6},
location = {Indianapolis, IN, USA}
}

@article{ABDI2024123988,
title = {Enhancing vessel arrival time prediction: A fusion-based deep learning approach},
journal = {Expert Systems with Applications},
volume = {252},
pages = {123988},
year = {2024},
issn = {0957-4174},
author = {Asad Abdi and Chintan Amrit},
}

@Article{app11104410,
AUTHOR = {Ogura, Takahiro and Inoue, Teppei and Uchihira, Naoshi},
TITLE = {Prediction of Arrival Time of Vessels Considering Future Weather Conditions},
JOURNAL = {Applied Sciences},
VOLUME = {11},
YEAR = {2021},
NUMBER = {10},
ARTICLE-NUMBER = {4410},
ISSN = {2076-3417},
}

@article{CHU2025105128,
title = {Vessel arrival time to port prediction via a stacked ensemble approach: Fusing port call records and AIS data},
journal = {Transportation Research Part C: Emerging Technologies},
volume = {176},
pages = {105128},
year = {2025},
issn = {0968-090X},
author = {Zhong Chu and Ran Yan and Shuaian Wang},
}

@misc{liu2024itransformerinvertedtransformerseffective,
      title={iTransformer: Inverted Transformers Are Effective for Time Series Forecasting}, 
      author={Yong Liu and Tengge Hu and Haoran Zhang and Haixu Wu and Shiyu Wang and Lintao Ma and Mingsheng Long},
      year={2024},
      eprint={2310.06625},
      archivePrefix={arXiv},
      primaryClass={cs.LG},
}

@inproceedings{Derrow_Pinion_2021, series={CIKM ’21},
   title={ETA Prediction with Graph Neural Networks in Google Maps},
   booktitle={Proceedings of the 30th ACM International Conference on Information and Knowledge Management},
   publisher={ACM},
   author={Derrow-Pinion, Austin and She, Jennifer and Wong, David and Lange, Oliver and Hester, Todd and Perez, Luis and Nunkesser, Marc and Lee, Seongjae and Guo, Xueying and Wiltshire, Brett and Battaglia, Peter W. and Gupta, Vishal and Li, Ang and Xu, Zhongwen and Sanchez-Gonzalez, Alvaro and Li, Yujia and Velickovic, Petar},
   year={2021},
   month=oct, pages={3767–3776},
   collection={CIKM ’21} 
}

@article{LEI2024116838,
title = {Predicting vessel arrival times on inland waterways: A tree-based stacking approach},
journal = {Ocean Engineering},
volume = {294},
pages = {116838},
year = {2024},
issn = {0029-8018},
author = {Jinyu Lei and Zhong Chu and Yong Wu and Xinglong Liu and Mingjun Luo and Wei He and Chenguang Liu},
}

@misc{vaswani2023attentionneed,
      title={Attention Is All You Need}, 
      author={Ashish Vaswani and Noam Shazeer and Niki Parmar and Jakob Uszkoreit and Llion Jones and Aidan N. Gomez and Lukasz Kaiser and Illia Polosukhin},
      year={2023},
      eprint={1706.03762},
      archivePrefix={arXiv},
      primaryClass={cs.CL},
}

@InProceedings{10.1007/978-3-031-43612-3_13,
author="Wenzel, Peter
and Jovanovic, Raka
and Schulte, Frederik",
title="A Neural Network Approach for ETA Prediction in Inland Waterway Transport",
booktitle="Computational Logistics",
year="2023",
publisher="Springer Nature Switzerland",
address="Cham",
pages="219--232",
isbn="978-3-031-43612-3"
}

@ARTICLE{8294051,
  author={Alessandrini, Alfredo and Mazzarella, Fabio and Vespe, Michele},
  journal={IEEE Transactions on Intelligent Transportation Systems}, 
  title={Estimated Time of Arrival Using Historical Vessel Tracking Data}, 
  year={2019},
  volume={20},
  number={1},
  pages={7-15},
}

@INPROCEEDINGS{10422495,
  author={Zhang, Xiaocai and Fu, Xiuju and Xiao, Zhe and Xu, Haiyan and Wei, Xiaoyang and Koh, Jimmy and Ogawa, Daichi and Qin, Zheng},
  booktitle={2023 IEEE 26th International Conference on Intelligent Transportation Systems (ITSC)}, 
  title={Prediction of Vessel Arrival Time to Pilotage Area Using Multi-Data Fusion and Deep Learning}, 
  year={2023},
  volume={},
  number={},
  pages={2268-2268},
}

@article{PARK2021100012, title = {Vessel estimated time of arrival prediction system based on a path-finding algorithm}, journal = {Maritime Transport Research}, volume = {2}, pages = {100012}, year = {2021}, issn = {2666-822X}, author = {Kikun Park and Sunghyun Sim and Hyerim Bae}, 
}

@article{bulk,
author = {El Mekkaoui, Sara and Benabbou, Loubna and Berrado, Abdelaziz},
year = {2022},
month = {10},
pages = {},
title = {Deep learning models for vessel’s ETA prediction: bulk ports perspective},
volume = {35},
journal = {Flexible Services and Manufacturing Journal},
}

@article{noman2025review,
  title={A Review of Vessel Time of Arrival Prediction on Waterway Networks: Current Trends, Open Issues, and Future Directions},
  author={Noman, Abdullah Al and Heuermann, Aaron and Wiesner, Stefan and Thoben, Klaus-Dieter},
  journal={Computers},
  volume={14},
  number={2},
  pages={41},
  year={2025},
  publisher={MDPI}
}

@misc{unctad,
  author = {UNCTAD},
  title = {Review of Maritime Transport 2024},
  howpublished ={\url{https://unctad.org/system/files/official-document/rmt2024_en.pdf}},
  year = 2024,
  note = "(Accessed 25 May 2025)"
}

@article{chu2024vessel,
  title={Are vessel arrival and port operations affected by COVID-19? Evidence from the Hong Kong port},
  author={Chu, Zhong and Yan, Ran and Wang, Shuaian},
  journal={Transport Policy},
  volume={154},
  pages={157--181},
  year={2024},
  publisher={Elsevier}
}

@article{wang2023innovative,
  title={Innovative approaches to addressing the tradeoff between interpretability and accuracy in ship fuel consumption prediction},
  author={Wang, Haoqing and Yan, Ran and Wang, Shuaian and Zhen, Lu},
  journal={Transportation Research Part C: Emerging Technologies},
  volume={157},
  pages={104361},
  year={2023},
  publisher={Elsevier}
}

@article{yang2024efficient,
  title={An efficient ranking-based data-driven model for ship inspection optimization},
  author={Yang, Ying and Yan, Ran and Wang, Shuaian},
  journal={Transportation Research Part C: Emerging Technologies},
  volume={165},
  pages={104731},
  year={2024},
  publisher={Elsevier}
}

@article{chu2024evaluation,
  title={Evaluation and prediction of punctuality of vessel arrival at port: a case study of Hong Kong},
  author={Chu, Zhong and Yan, Ran and Wang, Shuaian},
  journal={Maritime Policy \& Management},
  volume={51},
  number={6},
  pages={1096--1124},
  year={2024},
  publisher={Taylor \& Francis}
}

@article{el2022machine,
  title={Machine learning models for efficient port terminal operations: Case of vessels’ arrival times prediction},
  author={El Mekkaoui, Sara and Benabbou, Loubna and Berrado, Abdelaziz},
  journal={IFAC-PapersOnLine},
  volume={55},
  number={10},
  pages={3172--3177},
  year={2022},
  publisher={Elsevier}
}

@article{kolley2023robust,
  title={Robust berth scheduling using machine learning for vessel arrival time prediction},
  author={Kolley, Lorenz and R{\"u}ckert, Nicolas and Kastner, Marvin and Jahn, Carlos and Fischer, Kathrin},
  journal={Flexible services and manufacturing journal},
  volume={35},
  number={1},
  pages={29--69},
  year={2023},
  publisher={Springer}
}

@article{servos2019travel,
  title={Travel time prediction in a multimodal freight transport relation using machine learning algorithms},
  author={Servos, Nikolaos and Liu, Xiaodi and Teucke, Michael and Freitag, Michael},
  journal={Logistics},
  volume={4},
  number={1},
  pages={1},
  year={2019},
  publisher={MDPI}
}

@article{yan2021emerging,
  title={Emerging approaches applied to maritime transport research: Past and future},
  author={Yan, Ran and Wang, Shuaian and Zhen, Lu and Laporte, Gilbert},
  journal={Communications in Transportation Research},
  volume={1},
  pages={100011},
  year={2021},
  publisher={Elsevier}
}

@inproceedings{liu2023uncertainty,
  title={Uncertainty-aware probabilistic travel time prediction for on-demand ride-hailing at didi},
  author={Liu, Hao and Jiang, Wenzhao and Liu, Shui and Chen, Xi},
  booktitle={Proceedings of the 29th ACM SIGKDD Conference on Knowledge Discovery and Data Mining},
  pages={4516--4526},
  year={2023}
}

@inproceedings{
yuan2024spatiotemporal,
title={Spatio-Temporal Few-Shot Learning via Diffusive Neural Network Generation},
author={Yuan Yuan and Chenyang Shao and Jingtao Ding and Depeng Jin and Yong Li},
booktitle={The Twelfth International Conference on Learning Representations},
year={2024},
}

@misc{miao2024unifiedreplaybasedcontinuouslearning,
      title={A Unified Replay-based Continuous Learning Framework for Spatio-Temporal Prediction on Streaming Data}, 
      author={Hao Miao and Yan Zhao and Chenjuan Guo and Bin Yang and Kai Zheng and Feiteng Huang and Jiandong Xie and Christian S. Jensen},
      year={2024},
      eprint={2404.14999},
      archivePrefix={arXiv},
      primaryClass={cs.DB},
}

@misc{wenzhe2025stgdpmvesseltrajectorypredictionspatiotemporal,
      title={STGDPM:Vessel Trajectory Prediction with Spatio-Temporal Graph Diffusion Probabilistic Model}, 
      author={Jin Wenzhe and Tang Haina and Zhang Xudong},
      year={2025},
      eprint={2503.08065},
      archivePrefix={arXiv},
      primaryClass={cs.AI},
}

@article{Lee2021DataDrivenAF,
  title={Data-Driven Analysis for Safe Ship Operation in Ports Using Quantile Regression Based on Generalized Additive Models and Deep Neural Network},
  author={Hyeong-Tak Lee and Hyun Yang and Ik-Soon Cho},
  journal={Sensors (Basel, Switzerland)},
  year={2021},
  volume={21},
}

@article{kwun2021prediction,
  title={Prediction of vessel arrival time using auto identification system data},
  author={Kwun, Hyeonho and Bae, Hyerim},
  journal={Int J Innov Comput Inf Control},
  volume={17},
  number={2},
  pages={725--734},
  year={2021}
}

@misc{wang2024timexerempoweringtransformerstime,
      title={TimeXer: Empowering Transformers for Time Series Forecasting with Exogenous Variables}, 
      author={Yuxuan Wang and Haixu Wu and Jiaxiang Dong and Guo Qin and Haoran Zhang and Yong Liu and Yunzhong Qiu and Jianmin Wang and Mingsheng Long},
      year={2024},
      eprint={2402.19072},
      archivePrefix={arXiv},
      primaryClass={cs.LG},
      url={https://arxiv.org/abs/2402.19072}, 
}

@article{WANG2025121873,
title = {Predicting estimated time of arrival for ships: A frequency-based approach considering met-ocean factors},
journal = {Ocean Engineering},
volume = {337},
pages = {121873},
year = {2025},
issn = {0029-8018},
doi = {https://doi.org/10.1016/j.oceaneng.2025.121873},
url = {https://www.sciencedirect.com/science/article/pii/S0029801825015793},
author = {Yiyang Wang and Xiaonan Zhang and Yuhan Guo}
}

@misc{wu2022autoformerdecompositiontransformersautocorrelation,
      title={Autoformer: Decomposition Transformers with Auto-Correlation for Long-Term Series Forecasting}, 
      author={Haixu Wu and Jiehui Xu and Jianmin Wang and Mingsheng Long},
      year={2022},
      eprint={2106.13008},
      archivePrefix={arXiv},
      primaryClass={cs.LG},
      url={https://arxiv.org/abs/2106.13008}, 
}

@misc{zhou2022fedformerfrequencyenhanceddecomposed,
      title={FEDformer: Frequency Enhanced Decomposed Transformer for Long-term Series Forecasting}, 
      author={Tian Zhou and Ziqing Ma and Qingsong Wen and Xue Wang and Liang Sun and Rong Jin},
      year={2022},
      eprint={2201.12740},
      archivePrefix={arXiv},
      primaryClass={cs.LG},
      url={https://arxiv.org/abs/2201.12740}, 
}

@article{karmelic2025liner,
  title={Liner schedule reliability problem: An empirical analysis of disruptions and recovery measures in container shipping},
  author={Karmeli{\'c}, Jakov and Jovi{\'c} Mihanovi{\'c}, Marija and Peri{\'c} Had{\v{z}}i{\'c}, Ana and Br{\v{c}}i{\'c}, David},
  journal={Logistics},
  volume={9},
  number={4},
  pages={149},
  year={2025},
  publisher={MDPI}
}

@article{okur2022schedule,
  title={Schedule reliability in liner shipping: A study on global shipping lines},
  author={Okur, {\.I}lknur Gizem and Tuna, Okan},
  journal={Pomorstvo},
  volume={36},
  number={2},
  pages={389--400},
  year={2022},
  publisher={Sveu{\v{c}}ili{\v{s}}te u Rijeci, Pomorski fakultet}
}


\end{document}